\documentclass[10pt,twocolumn,letterpaper,pagenumbers]{article}

 \usepackage{cvpr}              
\usepackage{xr}
\usepackage{float}
\definecolor{cvprblue}{rgb}{0.21,0.49,0.74}
\usepackage[pagebackref,breaklinks,colorlinks,allcolors=cvprblue]{hyperref}

\def\paperID{23} 
\def\confName{CVPR}
\def\confYear{2025}

\title{How many classes do we need to see for novel class discovery?}

\author{Akanksha Sarkar\\
Cornell University\\
{\tt\small as2637@cornell.edu}
\and
Been Kim\\
Google DeepMind\\
{\tt\small beenkim@google.com}
\and
Jennifer Sun\\
Cornell University\\
{\tt\small jjs533@cornell.edu}
}

\begin{document}

\maketitle
\section{Introduction}
\label{sec:intro}
New classes, beyond what the model has seen during training, are frequently encountered in real-world applications, such as scientific discovery~\cite{segalin2021mouse,luxem2022identifying} or open world interactions~\cite{gao2023opengcd, liu2023open}. 
While machine learning datasets and use cases are rapidly evolving in terms of size, categories, and dimensions~\cite{kaplan2020scaling,deng2009imagenet,sun2020scalability}, we lack fundamental understanding of how neural networks effectively discover and learn new classes. To address this, our work focuses on General Class Discovery (GCD)~\cite{vaze2022generalized} and Novel Class Discovery (NCD)~\cite{troisemaine2023novel} in a controlled visual setting. 
\subsection{Introducing NCD and GCD}
GCD and NCD aims to identify and categorize novel classes in unlabeled data~\cite{vaze2022generalized,troisemaine2023novel}, building upon knowledge from a limited set of known classes.  
The setup is as follows \cite{troisemaine2023novel, hsu2017learning}: The model is provided with two distinct datasets: labeled set $\mathcal{D}_l$ and unlabeled set $\mathcal{D}_u$. The goal of GCD is to discover novel classes $C_u$ present in $\mathcal{D}_u$, and classify the known $C_l$ in $\mathcal{D}_l$. A sub-problem GCD is called the NCD problem which has the same setup with $\mathcal{D}_l$ (labeled) and $\mathcal{D}_u$ (unlabeled) with the goal to discover novel classes $C_u$ in $\mathcal{D}_u$, without doing any classifications on $D_l$.

\begin{figure*}[h]
  \centering
  \begin{subfigure}{0.9\linewidth}
  \includegraphics[width=1.05\textwidth]{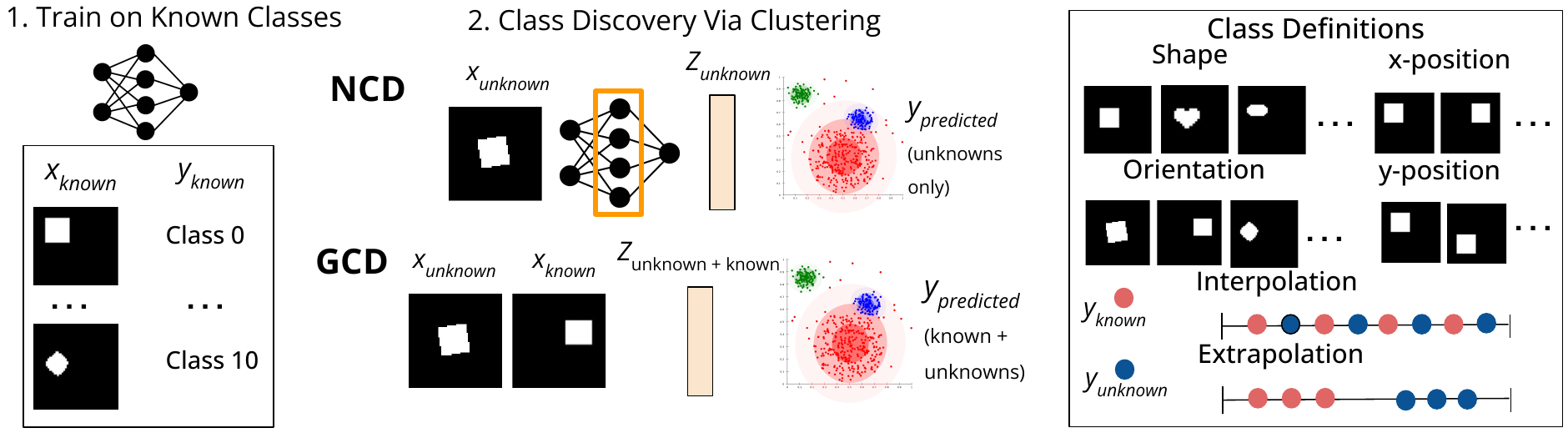}  
    \label{fig:short-a}
  \end{subfigure}
  \hfill
  \caption{\small We empirically study factors that influence success and failure of NCD and GCD. In both cases, we take a classical approach by first training a network on a set of known classes and using a clustering-based approach for class discovery. NCD focuses on labelling unknown classes only, while GCD is on both unknown and known classes. We consider factors of variation from dSprites~\cite{dsprites17} with more than 10 values; we augment the shape factor by creating new shapes between squares and circles. Finally, we evaluate class discovery for both interpolation from known classes and extrapolation. \vspace{-0.2in}}
  \label{fig:short}
\end{figure*}

\subsection{Clustering Algorithms for NCD.}
    The work by Hsu et al \cite{hsu2017learning} in 2018 first addresses the NCD problem by posing the problem as a transfer learning task where the unavailable labels of the target set need to be inferred. Clustering is a commonly used approach \cite{han2021autonovel, han2019learning, hsu2019multi, colin2024practical, fini2021unified} and serves as strong baselines in NCD. \textit{K-means} is widely used clustering algorithm among them with a lot of the methods either using \textit{k-means} \cite{han2021autonovel, hsu2019multi, colin2024practical} or other methods built on top of \textit{k-means} \cite{han2019learning, hsu2019multi}. 
    However, few study is done to disentangle what determines success and failure of class discovery mainly due to difficulty in working with real data. For instance, in behavioral neuroscience, a model learning to categorize mouse behaviors from video recordings might be influenced by factors like camera angle, cage lighting, or individual differences which are difficult to control~\cite{luxem2023open}. Furthermore, the {\it true} number of distinct behaviors exhibited by the mice are ambiguous even to expert observers~\cite{segalin2021mouse,anderson2014toward}, making it difficult to evaluate the model's performance. This calls for a controlled environment where well-defined factors of variation and ground truth labels. 

    \subsection{Probing Neural Networks using synthetic data.}
    The dSprites dataset \cite{dsprites17} is a synthetic dataset which has been used for various purposes including generative model interpretability \cite{liu2020towards} and detecting distribution shifts \cite{olson2021unsupervised, wijaya2021failing}. The dataset contains 2D shapes generated from 5 ground truth independent factors. These factors are shape, scale, orientation, x-position and y-position of a sprite. All the parameters except for shape are continuous in nature. Due to this we created a separate synthetic dataset which interpolates from a square to a circle in order to conduct experiments on a continuous scale for shape. 
    This dataset offers a simple visual environment where factors of variation (shape, orientation, position, etc.) are precisely controlled, enabling us to systematically investigate the impact of different factors on GCD and NCD performance. Specifically, we investigate how the number of training and novel classes, the nature of class-defining features, and the coverage of known classes (e.g., is the new class combination of existing classes or brand new?)  affect discovery. This is crucial in settings such as scientific discovery, where annotating more data (e.g., labeling more mouse behaviors) is possible, but time-consuming and expensive. We also generalize some of the observations from the synthetic dataset to real-life datasets. A better understanding of which behaviors to annotate and how many are necessary for effective discovery would enable scientists to make more informed choices. By leveraging the controlled nature of dSprites, we aim to gain simple yet deeper understanding of these fundamental questions in class discovery.
\vspace{-0.15in}
\begin{table}[ht]
\small 
\centering
\begin{tabular}{|c|c|c|c|}
\hline
\textbf{Exp. } & \textbf{Shape} & \textbf{Other Factors(x, y, orientation)} & \textbf{\#} \\
\hline
Exp A & fixed & 1 class-defining, 2 variable & 3 \\
Exp B & discrete & 1 class-defining, 1 variable, 1 fixed & 6 \\
Exp C & continuous & 2 variable, orientation fixed  & 3 \\
\hline
\end{tabular}
\caption{\small Summary of setup, based on whether shape is fixed, discrete, or continuous (interpolated between square and circle). Other factors are either class-defining (labels based on the variable), fixed, or variable. There is no orientation factor for the square/circle dataset in Exp C. The number of experiments (last column) corresponds to variations possible. For example, in Exp A, we have 3 experiments where class is either defined based on orientation, x-position, or y-position. \vspace{-0.1in}}
\label{tab:experiments}
\end{table}


\section{Experiments}
\label{sec:formatting}

\subsection{Experimental Setup.} We study the relationship between the number of known/unknown classes with class discovery performance. In each of our experiments, there is a class-defining factor, and other factors are either fixed or variable (Table~\ref{fig:example}). We then vary the number of known classes (i.e. the classes the model is trained on) and unknown classes (i.e. the classes the model is to discover). Each class is balanced and the same size. We show one experiment in Figure~\ref{fig:individual}, with additional experiments in appendix.

{ \bf Methodology.} We use a simple method based on the learned internal activations of neural network to cluster
for GCD and NCD  \cite{vaze2022generalized, troisemaine2023novel}, with the following steps: (1) feature extraction (activations), where the model is trained on the known classes and used as a feature extractor for unlabelled data, and (2) Clustering, where we apply an unsupervised clustering algorithm (K-means) to the extracted features to partition the data into distinct clusters corresponding to classes. Each cluster is then assigned a label based on the most frequent label in that cluster, with ties broken at random. The clustering methods commonly assumes the number of unknown classes is known a priori~\cite{troisemaine2023novel}. 

\begin{figure}[h]
\centering
\includegraphics[scale=0.4]{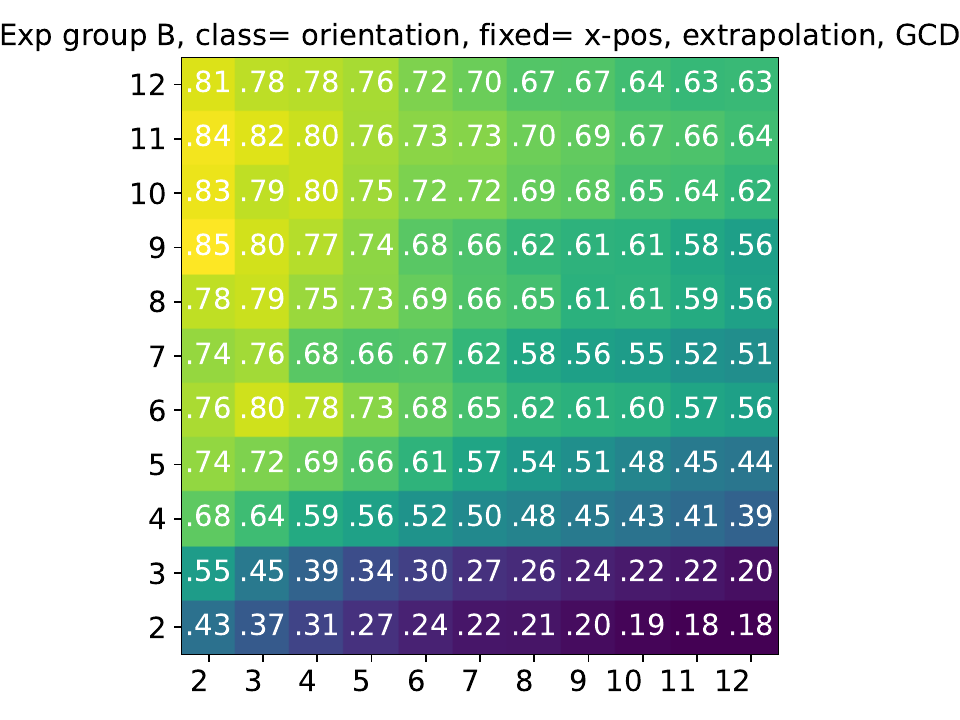}
\caption{\small Results from one experiment with row as number of unknowns and column as number of knowns.}\label{fig:individual}
\end{figure}

Here, we use a ResNet-18~\cite{he2016deep} pre-trained on ImageNet~\cite{deng2009imagenet} as the starting point for each experiment, and the model is fine-tuned to convergence on the known classes. We evaluate class discovery using accuracy, and discuss additional metrics (precision) in the appendix. To obtain summary metrics, we average across experiment groups in Table~\ref{fig:example}, and additional results for each experiment are in the appendix. We also perform the same set of experiments with different ResNet architectures, namely  ResNet-18 and ResNet-50, which differ in depth, model capacity, and computational complexity. This allows us to analyze the impact of model capacity on the performance and robustness.

\subsection{Testing with real Dataset.} 
The CUB-200 dataset~\cite{wah2011caltech} was used for NCD experiments. The dataset contains 200 bird species, organized into families and orders, enabling the design of both interpolation and extrapolation experiments. In interpolation setting, known and unknown classes are sampled from the same families and in extrapolation setting, known and unknown classes are selected from different families or orders.


\section{Results}

\subsection{Experiments with Synthetic Dataset}

Since the dataset is perfectly balanced, recall and accuracy values are the same. The plots shown only show precision and accuracy. 

\def\figsize{0.3}
\def\fighspace{-2mm}
\def\fighspacer{-2mm}
\begin{figure*}[h]
\centering
\begin{tabular}{cc}
\centering
\hspace{\fighspace}\includegraphics[width=\figsize\textwidth]{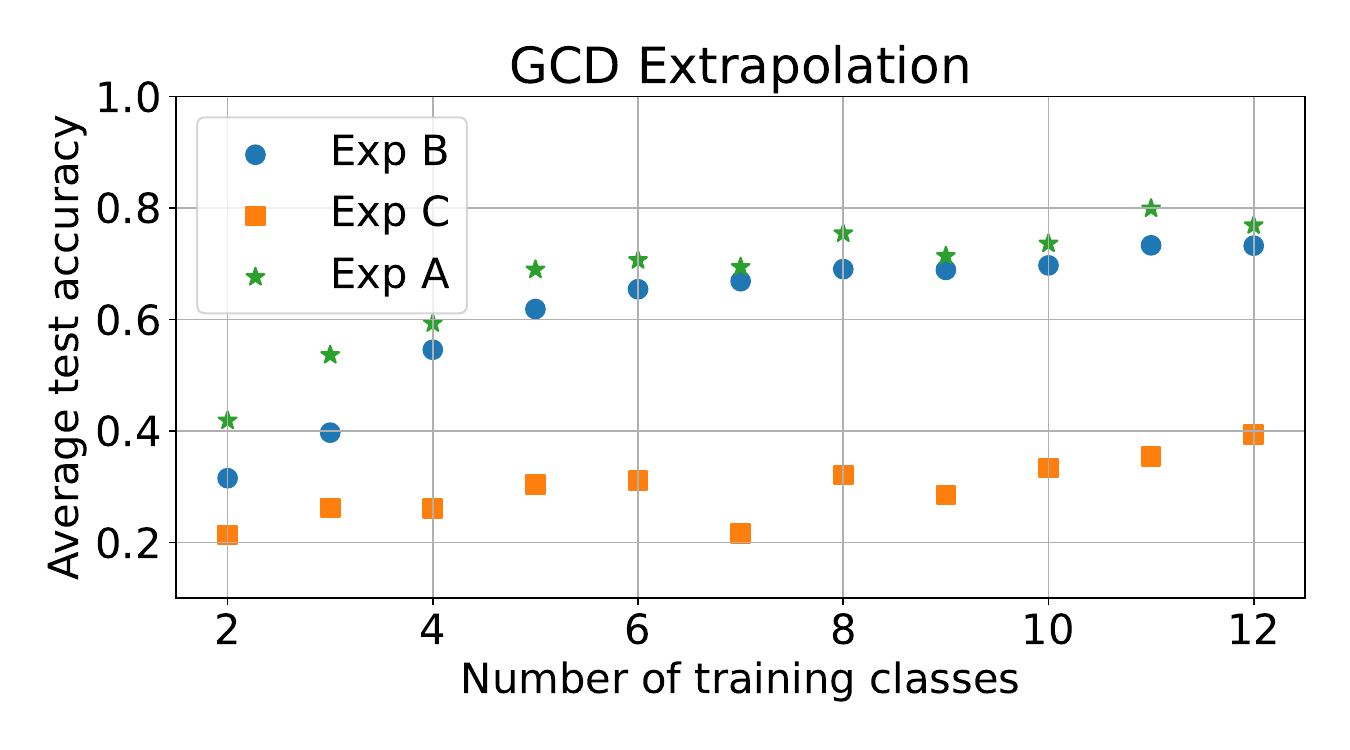}\hspace{\fighspace} & \includegraphics[width=\figsize\textwidth]{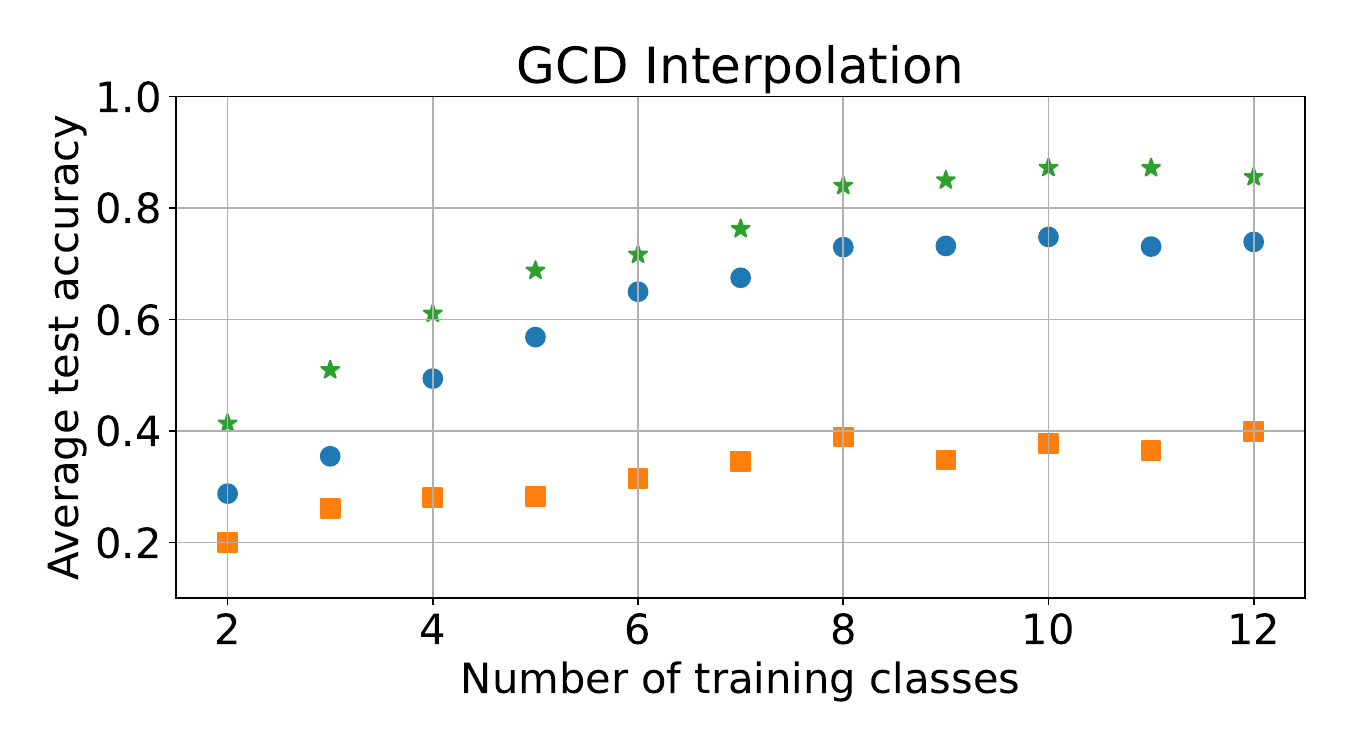}\hspace{\fighspacer}  \\
\hspace{\fighspace}\includegraphics[width=\figsize\textwidth]{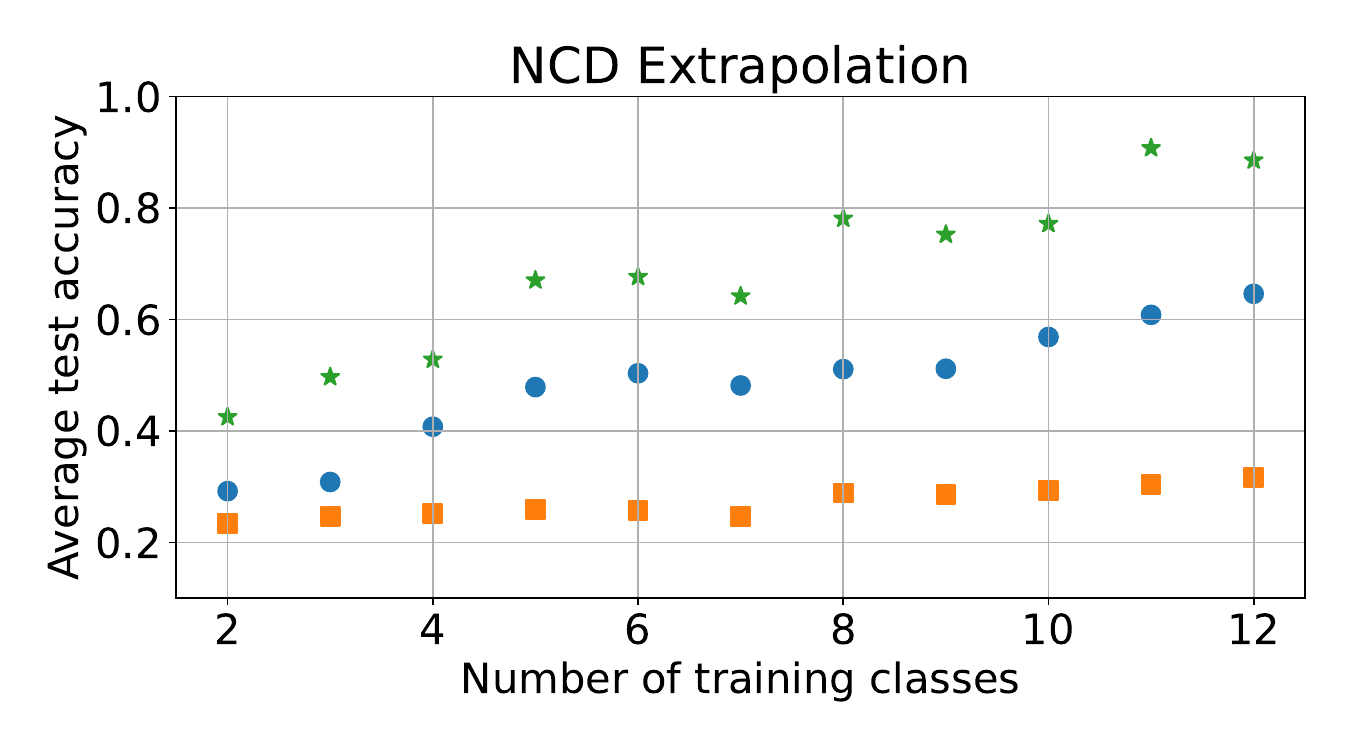}\hspace{\fighspace} & \includegraphics[width=\figsize\textwidth]{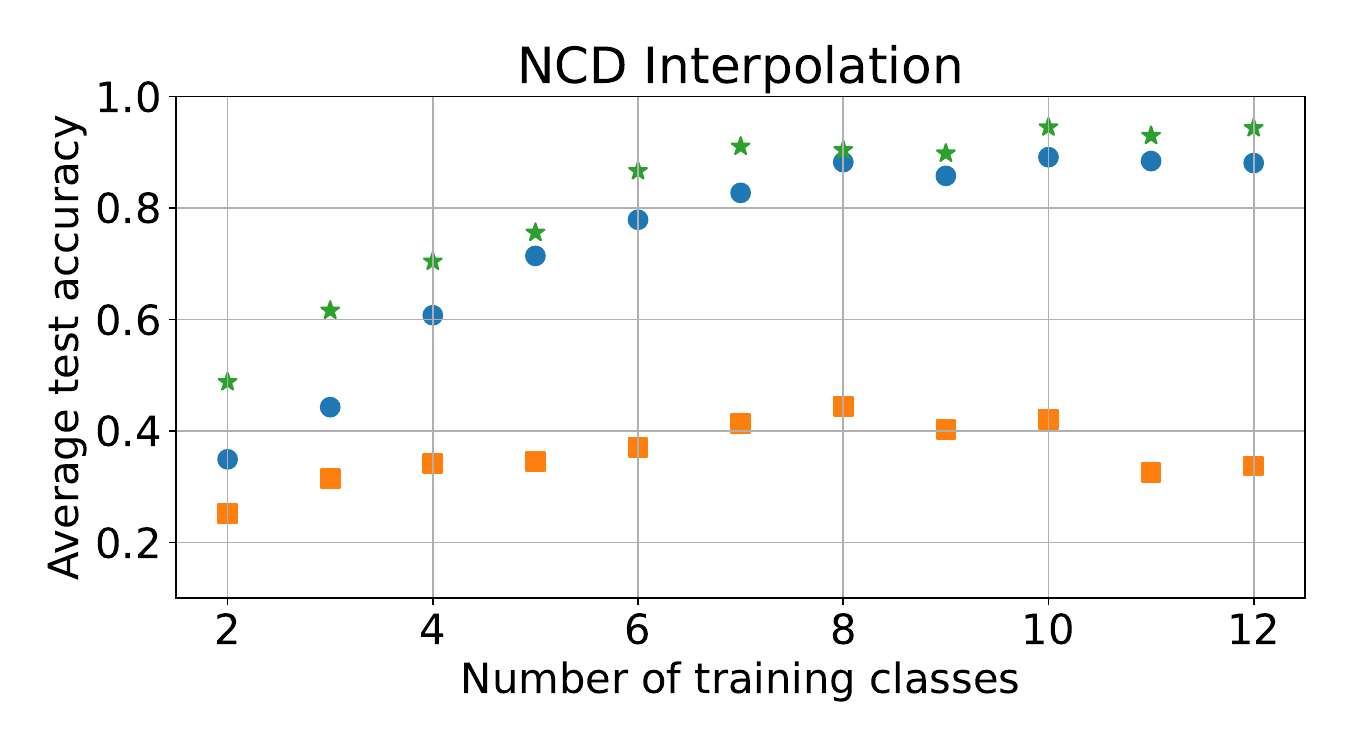}\hspace{\fighspacer} \\
\end{tabular}
\caption{\small Relationship between the number of training classes and average test accuracy, for GCD and NCD with extrapolation and interpolation. We calculate the average accuracy by averaging over all test classes for each experiment group.}
\label{fig:results}
\end{figure*}

\textbf{Relationship between number of known/unknown classes for discovery.}
In general, the accuracy naturally improves across all settings when the number of training classes increase (Figure~\ref{fig:results}). This is generally true across all the experiment groups (A, B, C), across GCD and NCD, as well as interpolation and extrapolation. 

Interestingly, we observe that when adding a new class, there is a more significant increase in accuracy when the amount of known classes is small (for Exp A and B). After a "saturation point" in the number of known classes, the increase in accuracy seems to plateau.  We note that for NCD interpolation (Figure~\ref{fig:results}), this number is around perfect class discovery, but this is not the case for the other experiments.
To further validate this observation, we performed an additional experiment to extend the number of training classes for Exp B, and find that this observation holds up to the maximum number of training classes for dSprites (see supplementary material).
We have not observed this trend in group C. This could be due to the relative difficulty of class discovery for the "shape" factor, and the experiment not yet reaching the saturation point.

{\it Interpolation and extrapolation.}
The above finding may make sense in hindsight; the model may have learned useful features for discovery after certain point. But what if the new class requires extrapolation--potentially learning new features? 
Naturally, the model performs better at interpolation compared to extrapolation, across all experiment groups. This difference is especially visible after the extrapolation point starts to plateau. However, both reaches the saturation point--an 
interesting indication that regardless of how 'new' the discovered classes are, there might be diminishing return on further labelling known classes. More rigorous study is required to bring this simple observation to an actionable insight for practitioners. 

{\it Exp A, B, C.} Across experiment groups, we find that the performance of A is followed by B, then C. This suggest that ``shape'' is a harder factor to learn, as the small difference between A and B is due to greater shape variability. In Exp C, where shape varies continuously between squares and circles, we observe that shape is harder to discover compared to the positional variations in Exp A and B.

\textbf{Relationship between Model-capacity and discovery} 

\begin{figure} [ht]
    \centering
    \includegraphics[width=0.5\textwidth]
    {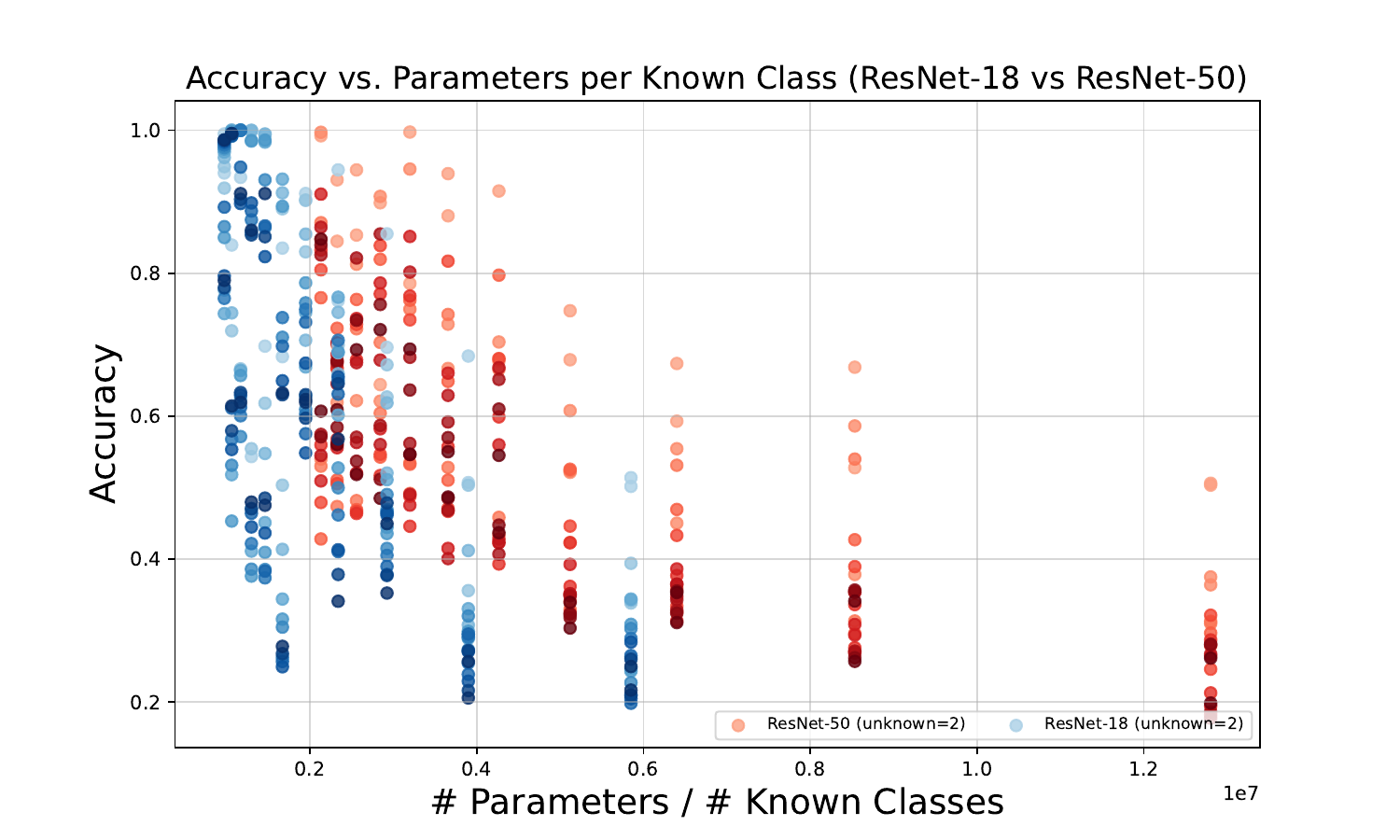}  
    \caption{\small Impact of model capacity (measured as the number of parameters divided by the number of known classes) on NCD accuracy for ResNet-18 and ResNet-50. The plots were made across Experiment A and B all of which exhibit similar trends.}
    \label{fig:mc_exp}
\end{figure}

As observed in Fig.~\ref{fig:mc_exp}, Both models exhibit a general trend of increasing accuracy as the number of parameters per known class increases. This suggests that larger model capacities relative to the number of known classes help in achieving better feature representation, leading to improved class discovery. However, the improvements tend to diminish at higher parameter-to-class ratios, suggesting that beyond a certain capacity, the additional model complexity does not significantly enhance discovery performance. This finding is consistent with the idea that the learning of novel class representations is constrained by factors such as feature diversity in the known classes and the inherent variability in the unknown classes.

There is no consistent trend where ResNet-50 significantly outperforms ResNet-18. In many cases, the performances overlap, particularly at lower parameter-to-class ratios. Therefore, increase in model capacity does not significantly help the discovery task. 

\subsubsection{Experiments with Real-Dataset}

\begin{figure} [h]
    \centering
    \includegraphics[width=0.5\textwidth]{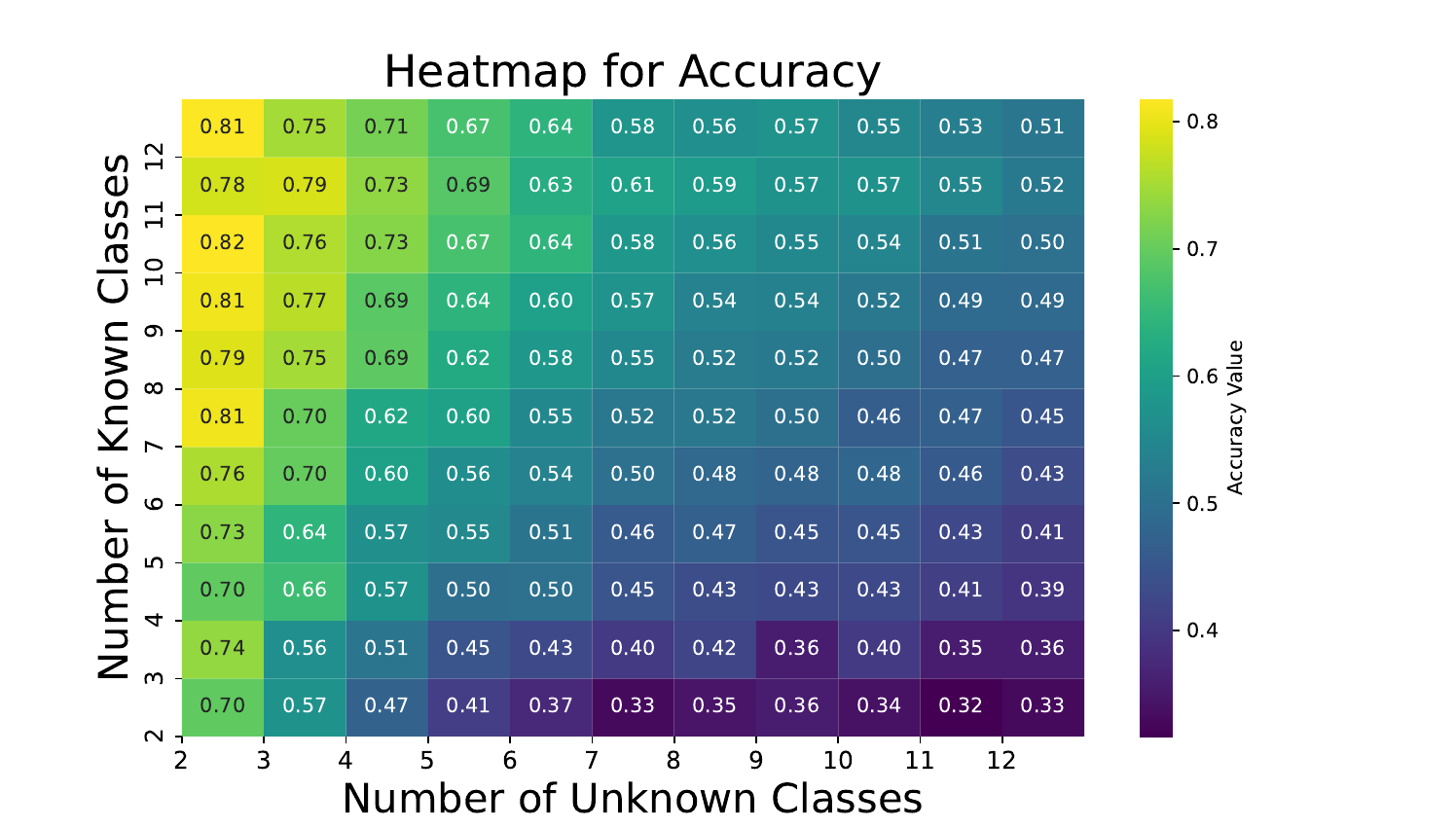} 
    \caption{\small Results from CUB-200 dataset experiment with row as number of known and column as unknown}
    \label{fig:example}
\end{figure}

The results on the CUB-200 dataset demonstrate how the trends observed in synthetic experiments generalize to real-world data. The heatmap for accuracy (Fig. ~\ref{fig:example}) reveal that performance varies depending on the number of known and unknown classes, with higher accuracy and precision generally achieved when the number of known classes is larger and the number of unknown classes is smaller. This is consistent with observations from synthetic datasets, where increased known class diversity improves feature representation, aiding in novel class discovery. However, the CUB-200 results also highlight the challenges posed by real-world variability, such as feature overlap and noise, which can impact precision and lead to reduced performance compared to synthetic settings. The gradual decrease in precision and accuracy with more unknown classes underscores the increasing complexity of discovering novel classes in fine-grained datasets like CUB-200, which exhibit high intra-class variability.

\textit{Interpolation and extrapolation}. Experiments around class coverage were also conducted using the real dataset. However, the results reveal no clear general trend in the difference between extrapolation and interpolation across precision and accuracy metrics. The variations in differences of the two sets of experiments appeared context-dependent and scattered across the range of known and unknown classes without a consistent pattern. While certain configurations, such as a higher number of known classes combined with fewer unknown classes, tend to show better extrapolation performance, this is not universally true. Similarly, interpolation does not consistently outperform extrapolation in scenarios with many unknown classes. These results suggest that the relative effectiveness of extrapolation versus interpolation is highly dependent on specific task configurations, indicating that neither approach is universally superior across all scenarios.

\section{Conclusion}
Our empirical results provide insight to  practitioners on factors that influence the success of class discovery. 
This is useful for the cost-benefit analysis in annotating new classes.
Our experiments indicate the performance gains from increasing the number of known classes plateau after a saturation point across multiple settings, suggesting a strategic limit to the resources invested in annotating new classes. 
We chose a synthetic dataset in our study for our simple controlled experimental framework, which serves as a starting point for future research of class discovery on more complex real-world datasets. We also try to generalize some of the observations to real-life datasets and ensure the realiability of results by testing across different model capacities.  
{
    \small
    \bibliographystyle{ieeenat_fullname}
    \bibliography{main}
}
\clearpage
\clearpage
\setcounter{page}{1}
\maketitlesupplementary

\section{Details of Dataset Used}

dSprites dataset has the following factors: shape (squares, hearts and ellipses), scale (ranging from 0 to 5), orientation (ranging from 0 to 39), x-position (ranging from 0 to 31) and y-position (ranging from 0 to 31). Since the dataset only contained limited options on the shape, in order to fully test this aspect, we extended the dataset by creating another synthetic dataset with the following properties: shape (ranging from 0 to 19), x-position (ranging from 0 to 9) and   y-position (ranging from 0 to 9). 

{\bf Continuous shape dataset.} The continuous shape dataset was generated by interpolating between square and circle shapes, while also varying their positions along the x and y axes. The square and circle shapes are defined as a set of points on a 64x64 canvas and interpolation is performed by using parameter $\alpha$ which ranges from 0 (square) to 1 (circle). For each shape in the dataset, the points are linearly interpolated between the square and circle, creating intermediate shapes (squircles). Additionally, for each shape, x and y positions are adjusted within defined shift ranges, producing variational positionings. The shapes generated are binary images. 

\begin{figure}[h]
    \centering
    \begin{minipage}{0.2\textwidth}
        \centering
        \includegraphics[width=\textwidth]{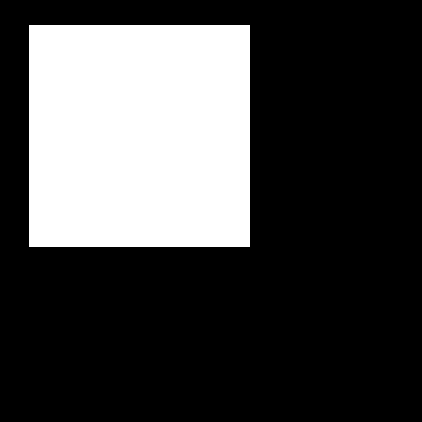}
        \label{fig:image1}
    \end{minipage}
    \begin{minipage}{0.2\textwidth}
        \centering
        \includegraphics[width=\textwidth]{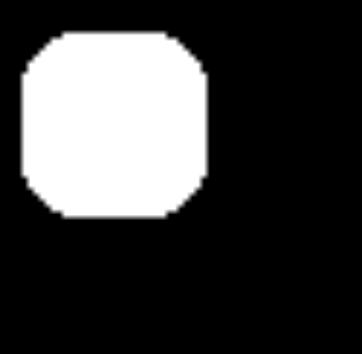}
        \label{fig:image2}
    \end{minipage}
    \begin{minipage}{0.2\textwidth}
        \centering
        \includegraphics[width=\textwidth]{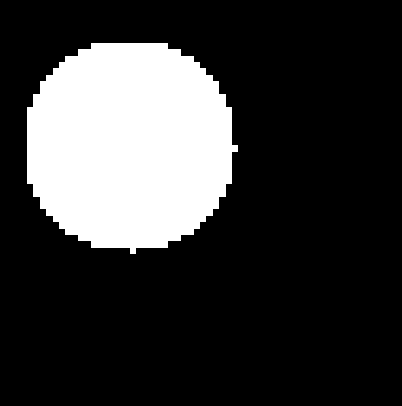}
        \label{fig:image3}
    \end{minipage}
    \caption{Showing interpolation from square to circle used for the continuous shape dataset.}
\end{figure}

\section{Evaluation Metrics}
In our experiments, we used accuracy and precision as evaluation metrics. Accuracy is calculated as the ratio of correctly predicted instances to the total number of instances, providing an overall measure of performance. Precision is calculated as the ratio of true positive predictions to the total number of positive predictions made by the model, focusing on the model's ability to correctly identify positive instances. Recall is calculated as the ratio of true positives to the actual number of positive instances, indicating the model’s ability to capture all relevant positive instances. Since the dataset is perfectly balanced, accuracy and recall are identical, as both metrics reflect the proportion of correctly predicted instances.

\section{Experiments with synthetic Dataset}

In the result section of the paper, we showcased the relationship between the number of training classes and average test accuracy, for GCD and NCD with extrapolation and interpolation. The average accuracy was calculated by averaging over all test classes for each experiment group. Here Fig.~\ref{tab:tableee} - ~\ref{tab:12} shown below, showcase the individual experiments done using synthetic dataset across experiment class A,B and C as described in the method section. These cover both, NCD as well as GCD experiments. (These form the experiments used to get the final graph in results).

\begin{figure}[H]
  \centering

  \includegraphics[width=\linewidth]{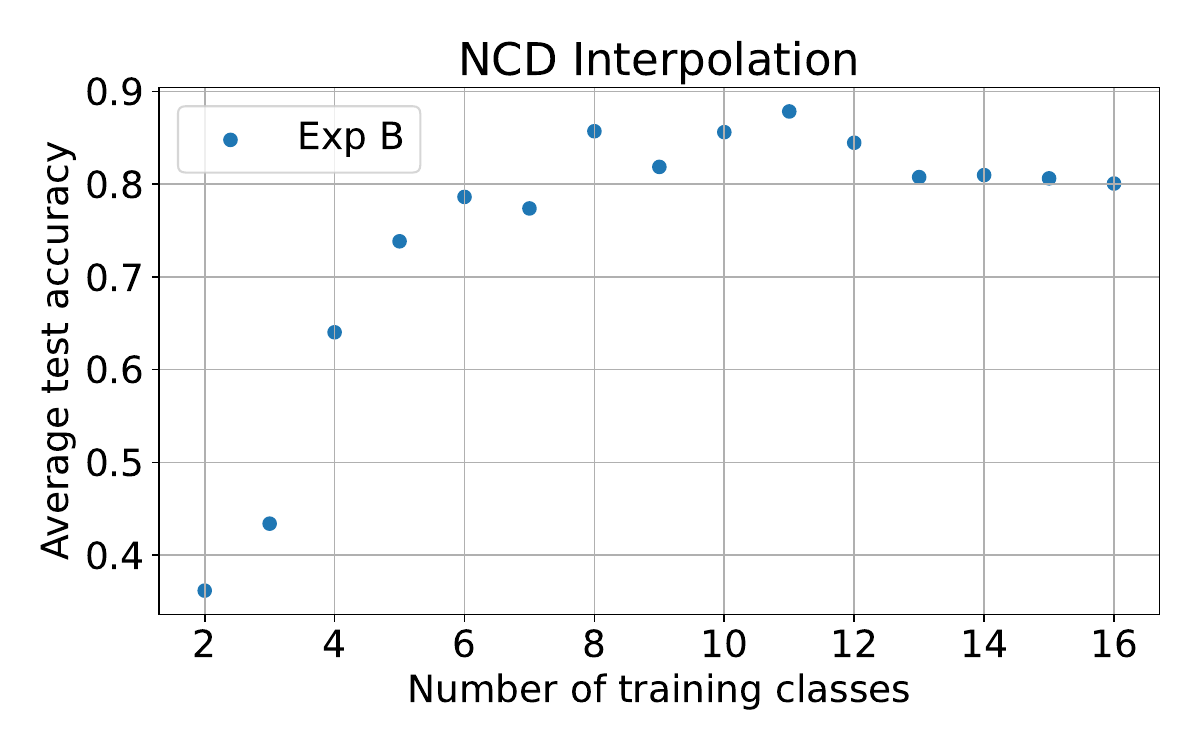}\par\vspace{0.4em}
  \includegraphics[width=\linewidth]{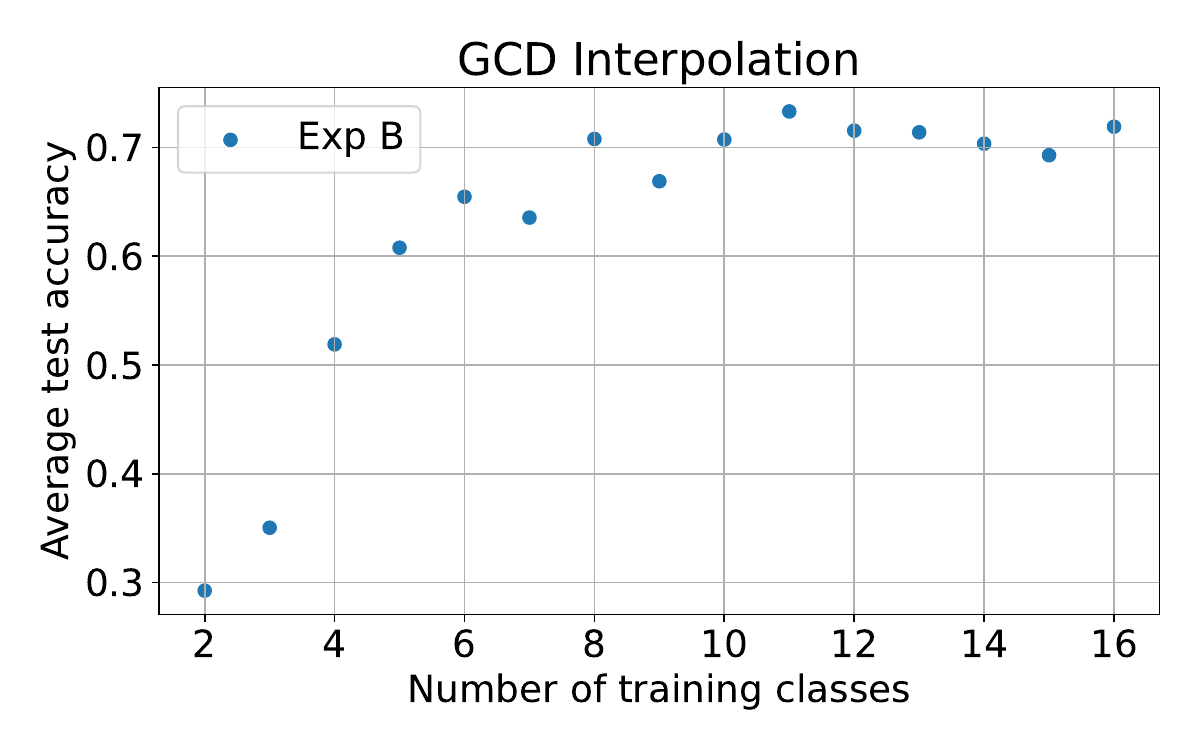}\par\vspace{0.4em}
  \includegraphics[width=\linewidth]{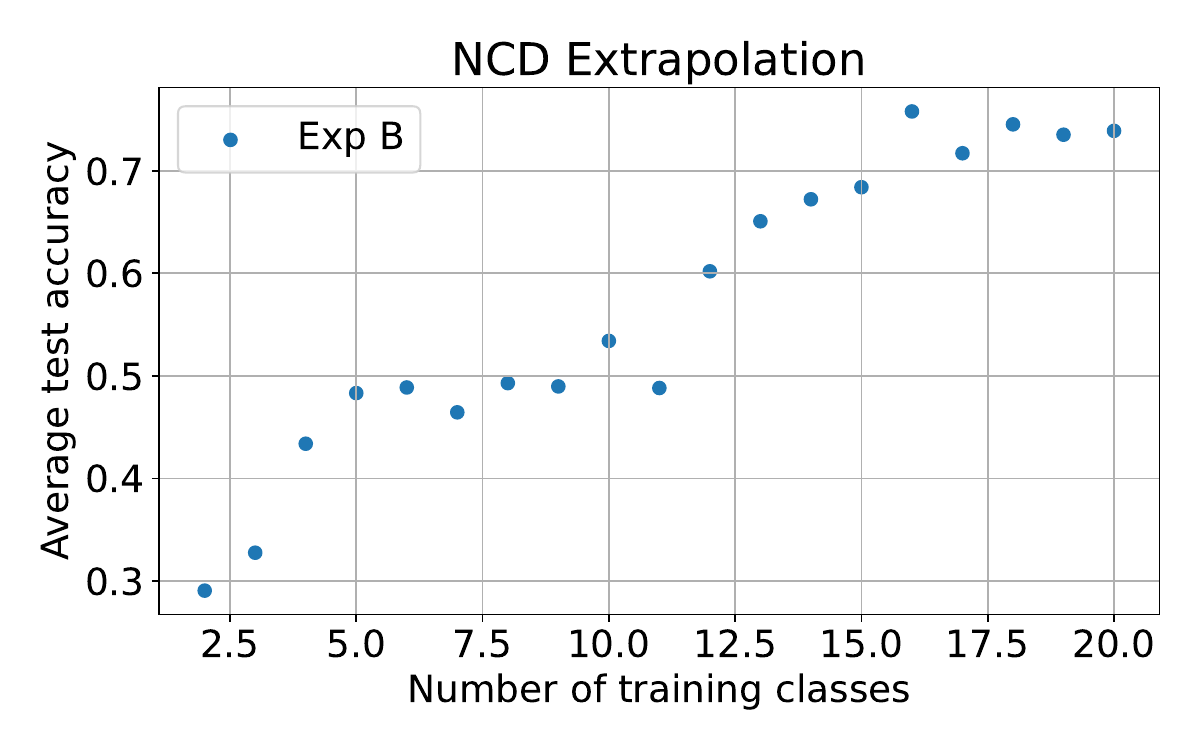}\par\vspace{0.4em}
  \includegraphics[width=\linewidth]{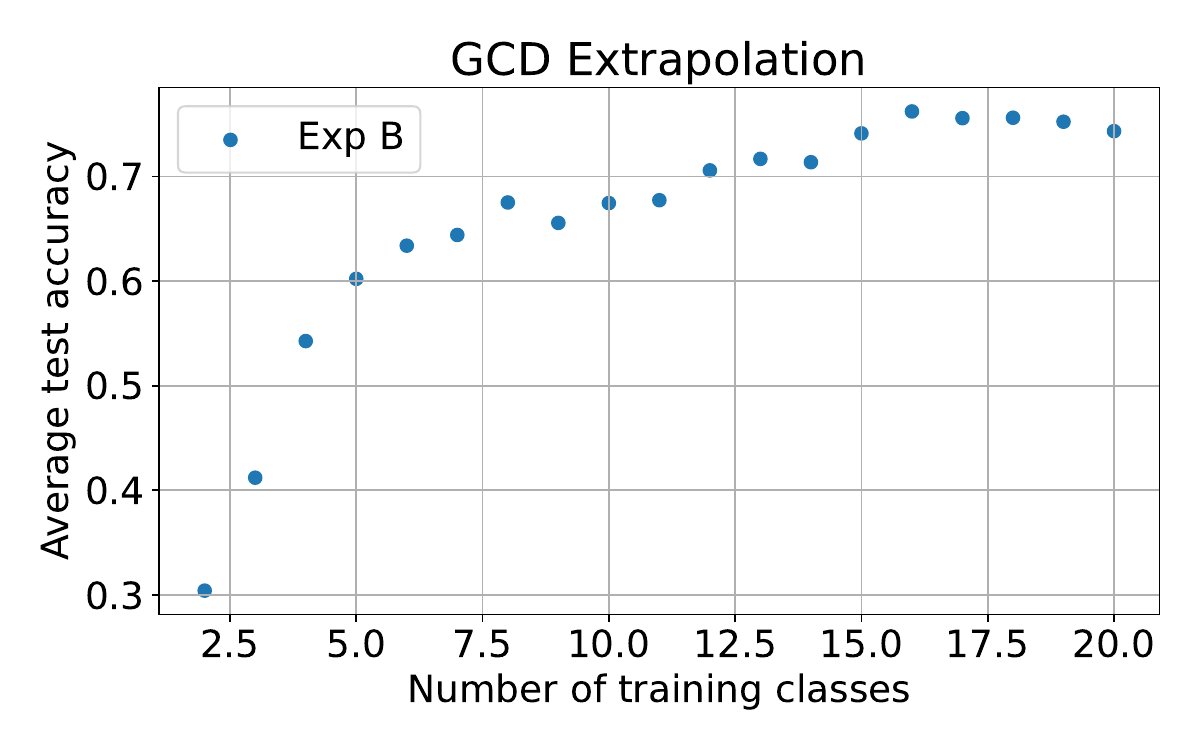}

  \caption{Results from training Exp B on more classes. Accuracy plateaus across settings.}
  \label{fig:graph_acc}
\end{figure}

Additionally, fig. ~\ref{fig:graph_acc} Showcases further testing our hypothesis of accuracy plateauing by testing Exp B on more classes. We notice that the expected trend holds.

\def\figsize{0.5}
\def\fighspace{-2mm}
\def\fighspacer{-2mm}

\begin{figure*}[p]
\centering
\begin{tabular}{cc}
\hspace{\fighspace}\includegraphics[width=\figsize\textwidth]{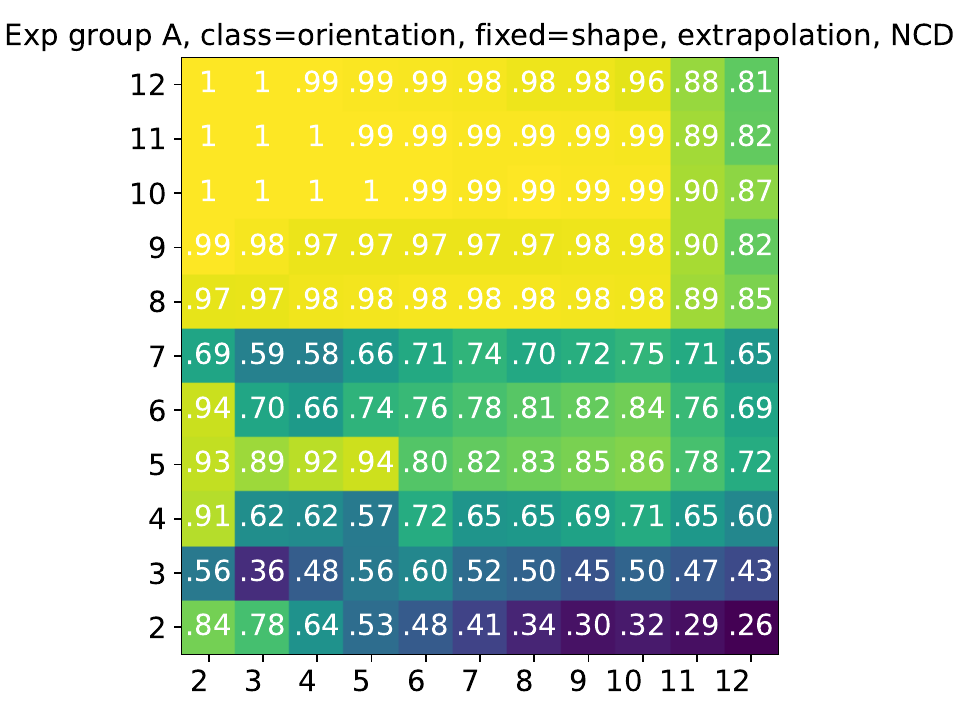}\hspace{\fighspace} &
\includegraphics[width=\figsize\textwidth]{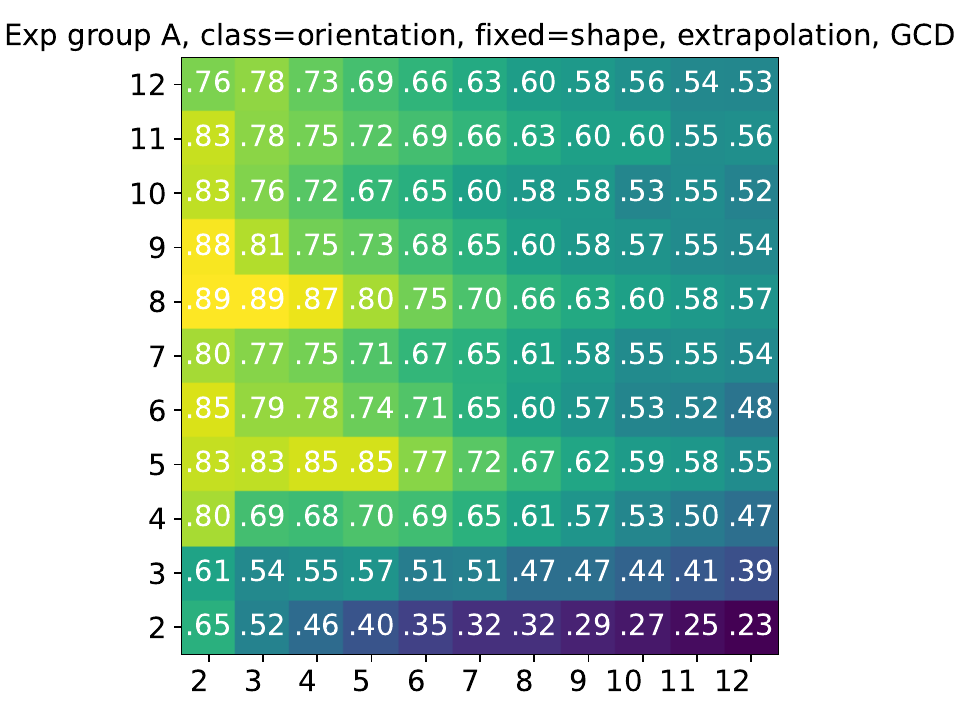}\hspace{\fighspacer} \\
\hspace{\fighspace}\includegraphics[width=\figsize\textwidth]{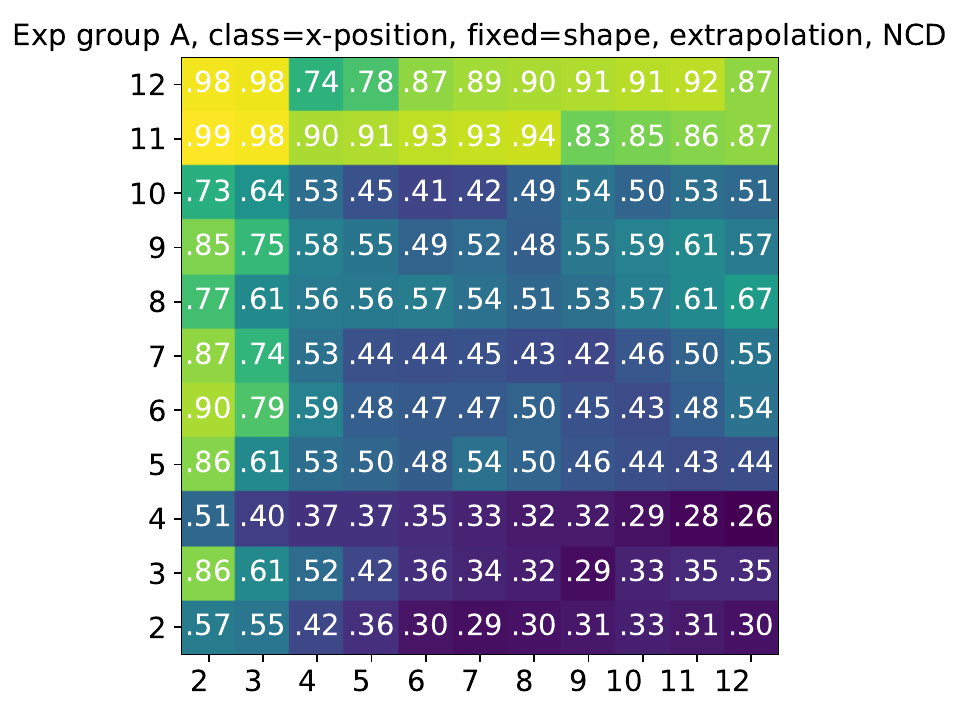}\hspace{\fighspace} &
\includegraphics[width=\figsize\textwidth]{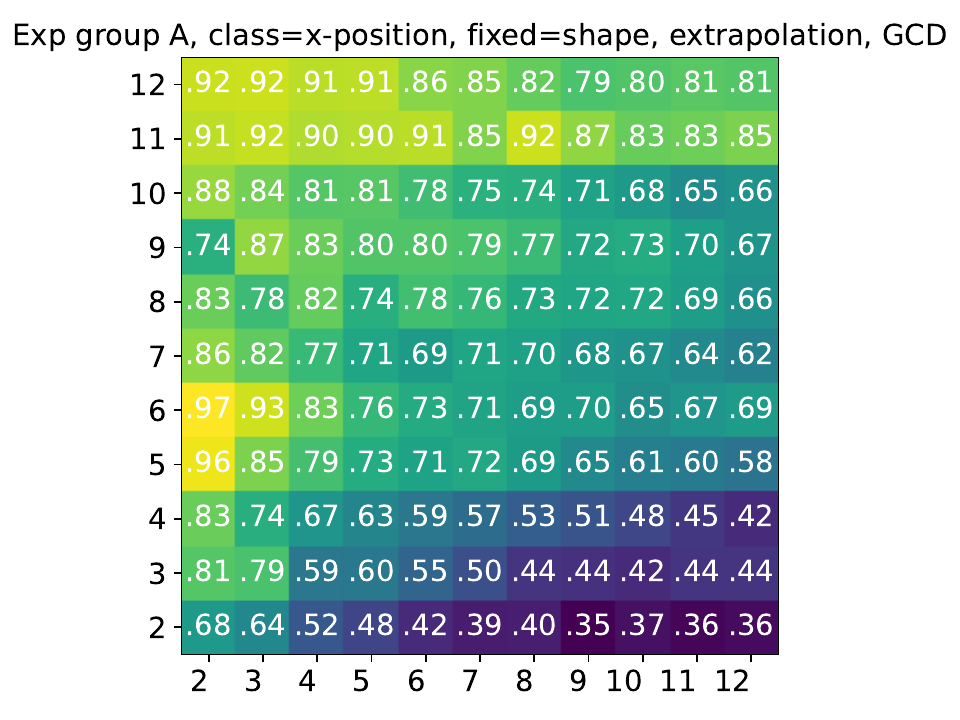}\hspace{\fighspacer} \\
\hspace{\fighspace}\includegraphics[width=\figsize\textwidth]{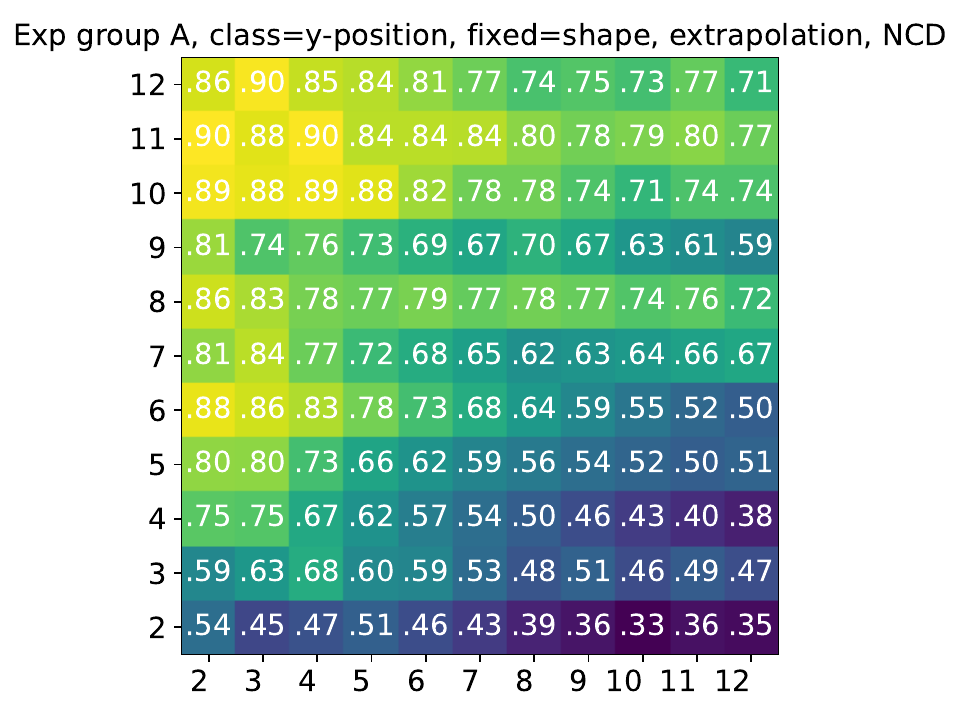}\hspace{\fighspace} &
\includegraphics[width=\figsize\textwidth]{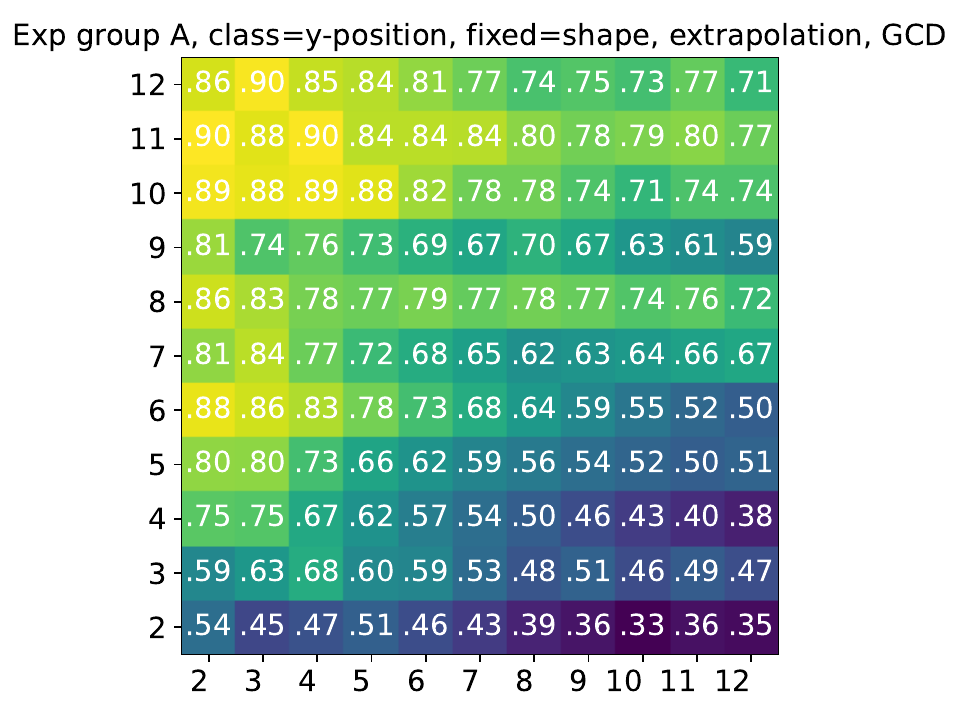}\hspace{\fighspacer} \\
\end{tabular}
\caption{\textbf{Extrapolation} results for Experiment group A: The heatmaps above are plots of the accuracy of discovering unseen classes (NCD) and classifying unseen and seen classes (GCD)}
\label{tab:tableee}
\end{figure*}

\def\figsize{0.5}
\def\fighspace{-2mm}
\def\fighspacer{-2mm}
\begin{figure*}[p]
\centering
\begin{tabular}{cc}
\centering
\hspace{\fighspace}

\includegraphics[width=\figsize\textwidth]{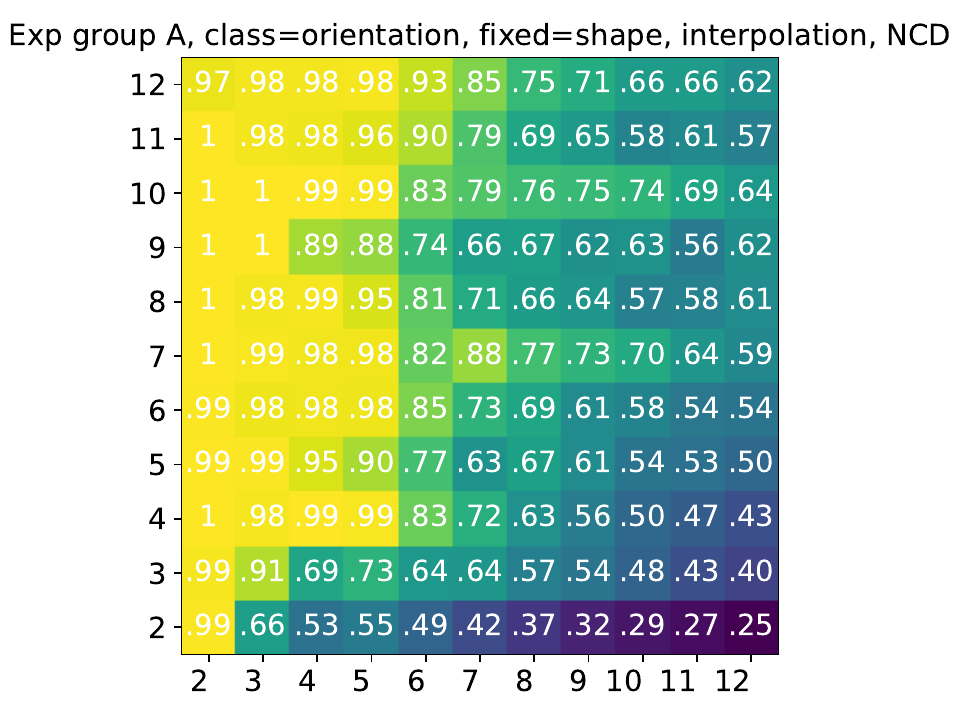} \hspace{\fighspace} & 

\includegraphics[width=\figsize\textwidth]{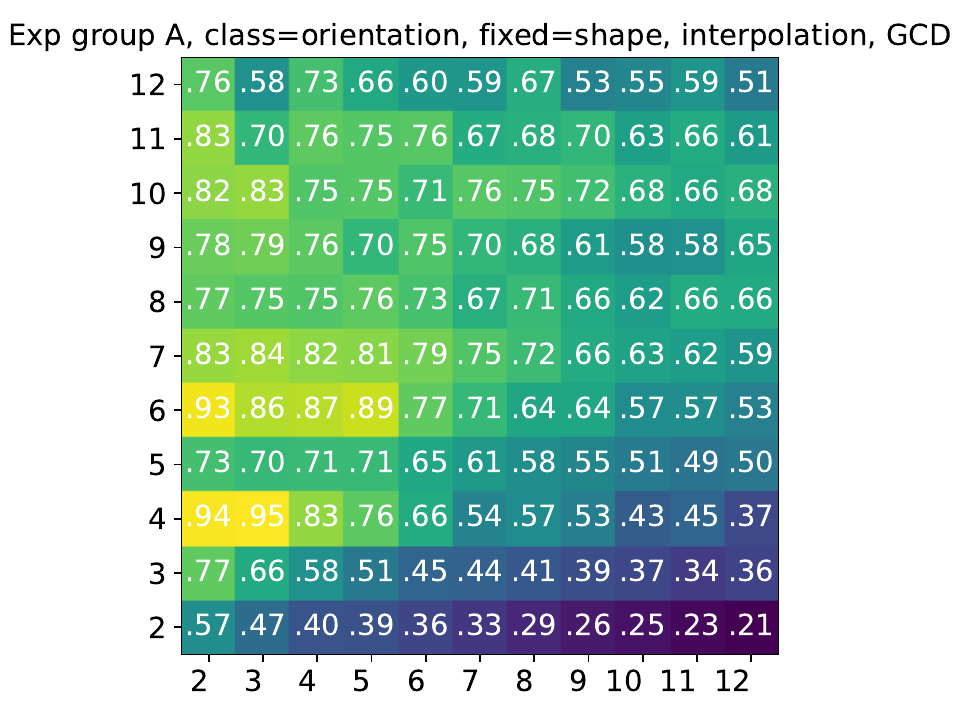}\hspace{\fighspacer}  \\

\hspace{\fighspace}\includegraphics[width=\figsize\textwidth]{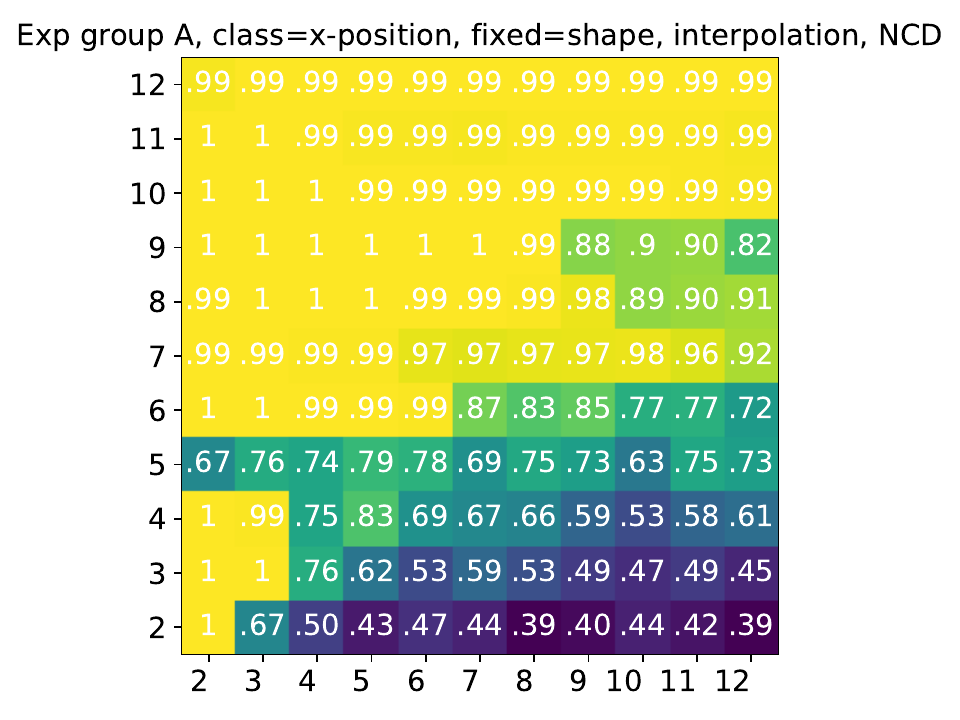}\hspace{\fighspace} & 

\includegraphics[width=\figsize\textwidth]{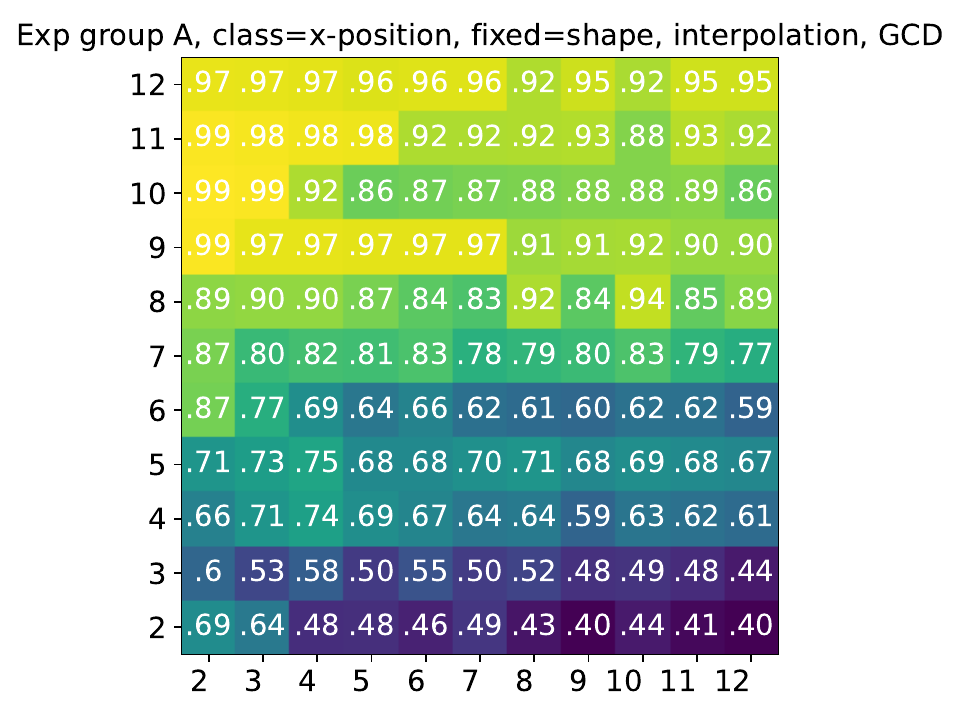}\hspace{\fighspacer} \\

\hspace{\fighspace}\includegraphics[width=\figsize\textwidth]{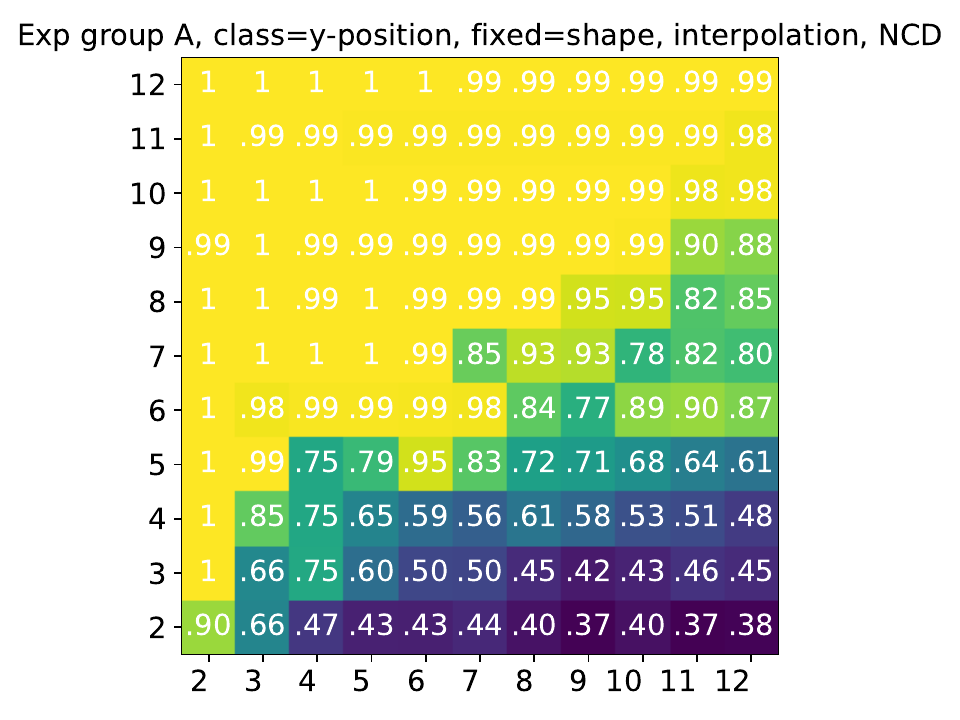}\hspace{\fighspace} & 

\includegraphics[width=\figsize\textwidth]{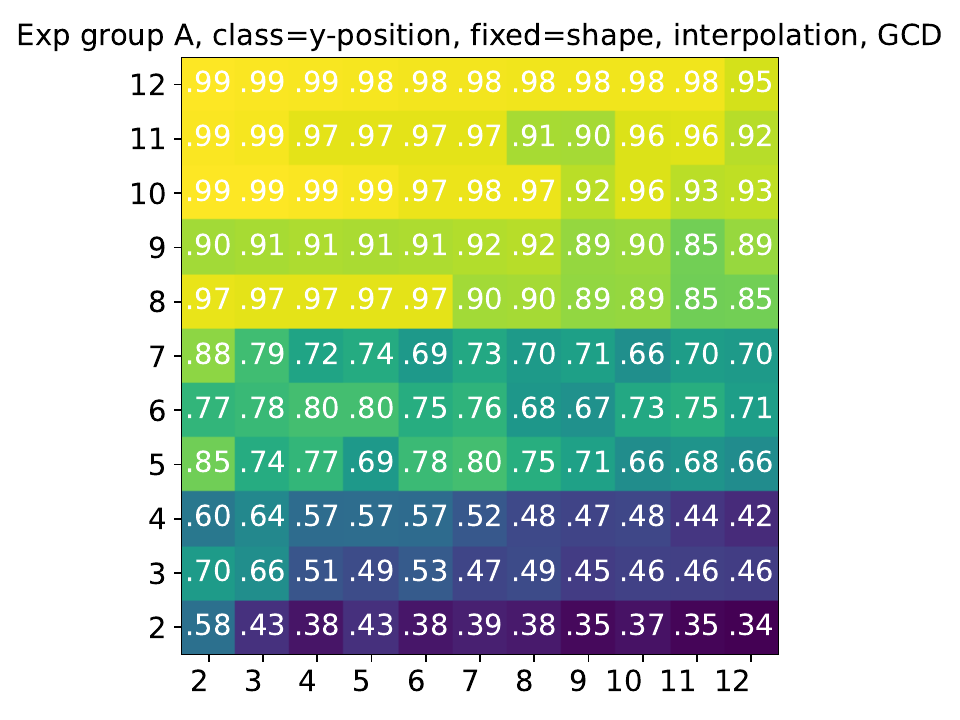}\hspace{\fighspacer} \\

\end{tabular}
\caption{ \textbf{Interpolation} results for Experiment group A : The heatmaps above are plots of the accuracy of discovering unseen classes (NCD) and classifying unseen and seen classes (GCD) }
\end{figure*}

\def\figsize{0.5}
\def\fighspace{-2mm}
\def\fighspacer{-2mm}
\begin{figure*}[p]
\centering
\begin{tabular}{cc}
\centering
\hspace{\fighspace}

\includegraphics[width=\figsize\textwidth]{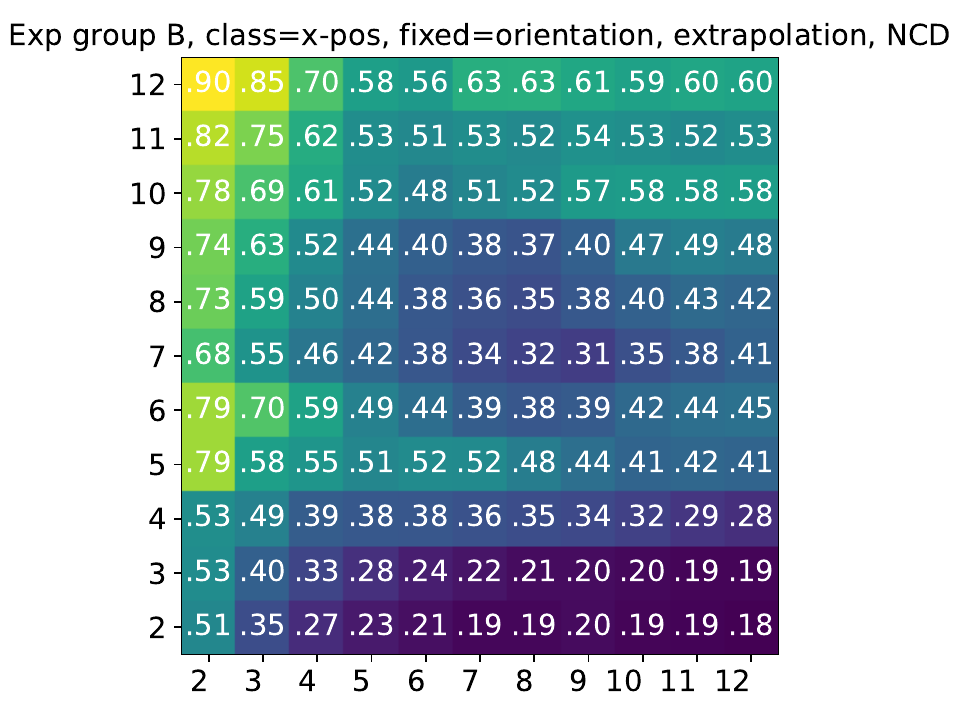} \hspace{\fighspace} & 

\includegraphics[width=\figsize\textwidth]{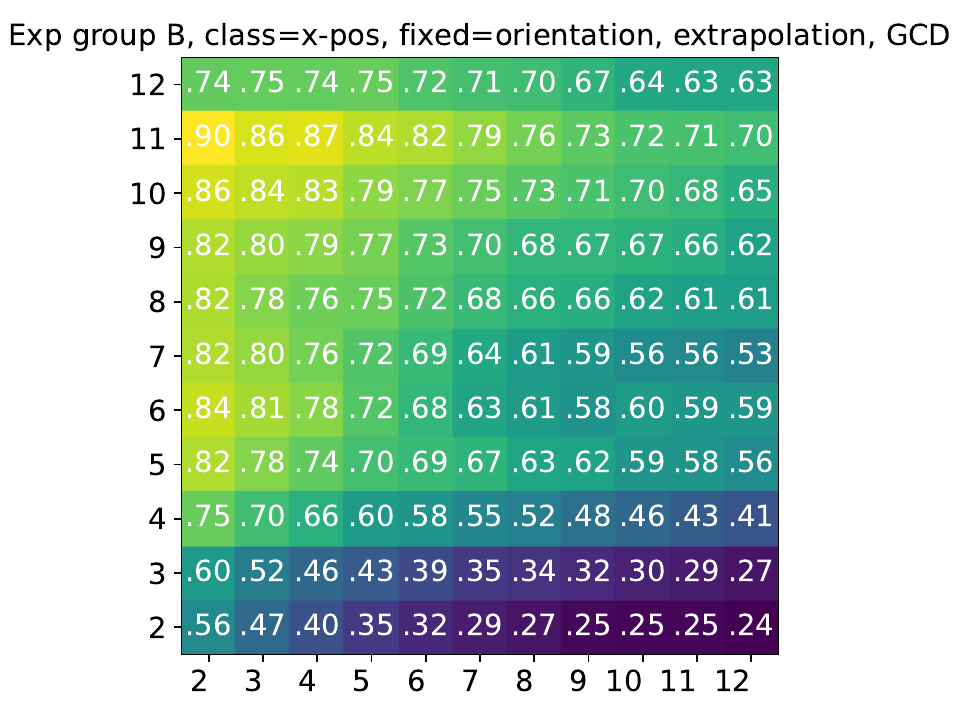}\hspace{\fighspacer}  \\

\hspace{\fighspace}\includegraphics[width=\figsize\textwidth]{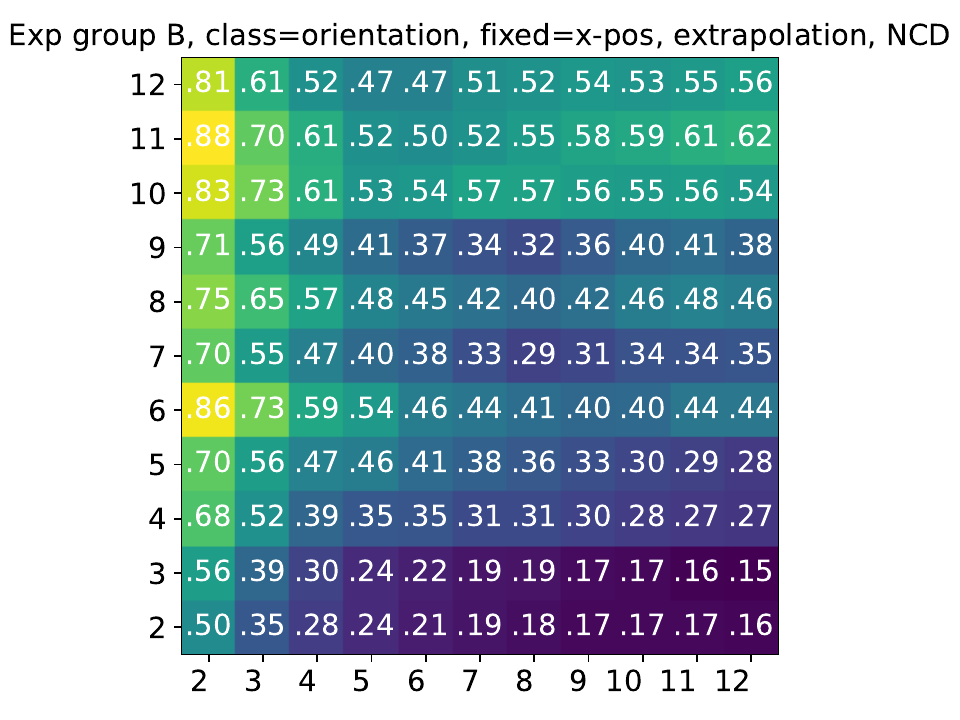}\hspace{\fighspace} & 

\includegraphics[width=\figsize\textwidth]{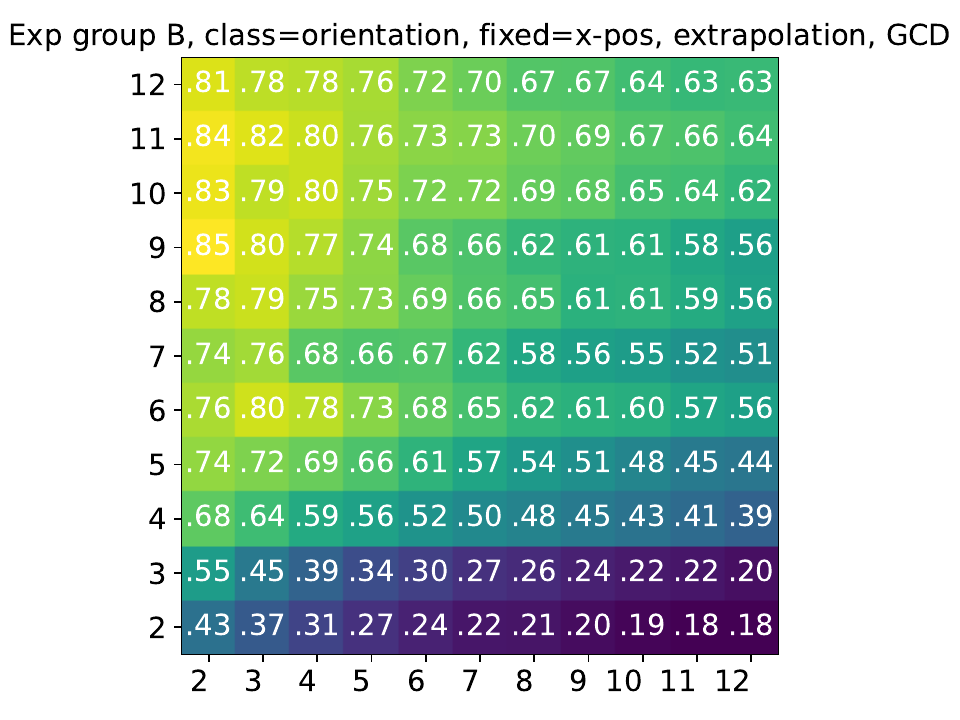}\hspace{\fighspacer} \\

\hspace{\fighspace}\includegraphics[width=\figsize\textwidth]{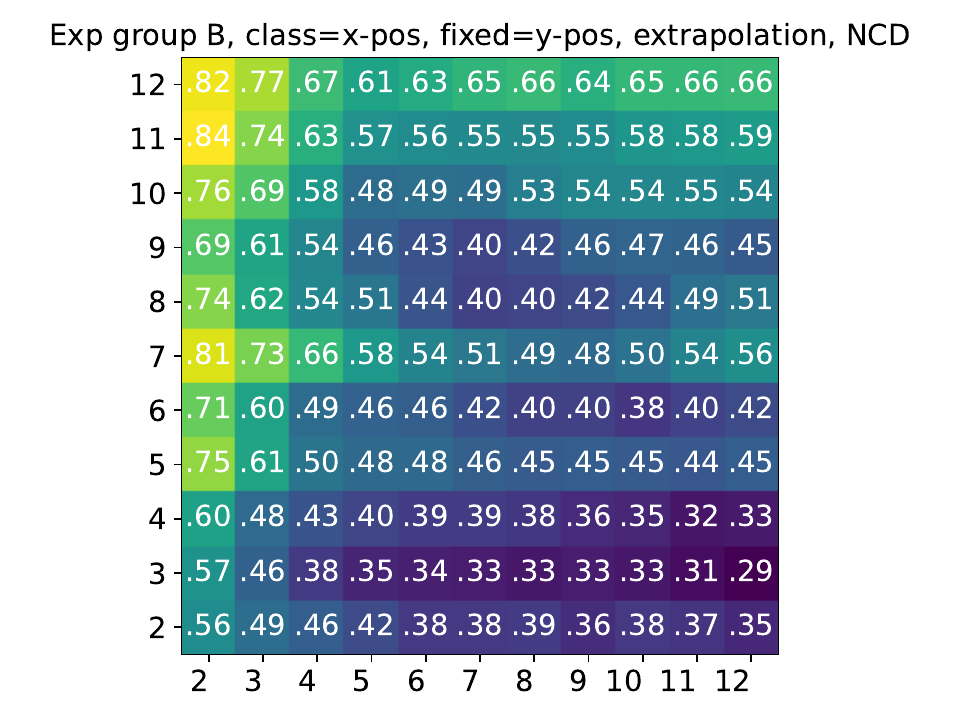}\hspace{\fighspace} & 

\includegraphics[width=\figsize\textwidth]{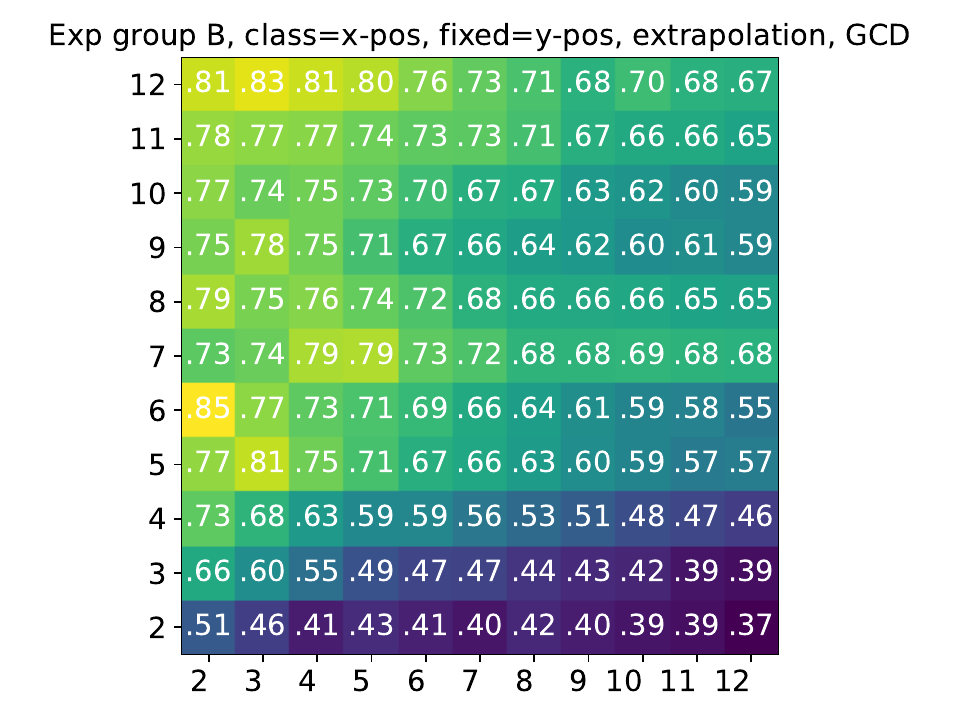}\hspace{\fighspacer} \\

\end{tabular}
\caption{ \textbf{Extrapolation} results for Experiment group B : The heatmaps above are plots of the accuracy of discovering unseen classes (NCD) and classifying unseen and seen classes (GCD). This is a subgroup of the plots chosen since the other plots show similar trends. }
\end{figure*}

\def\figsize{0.5}
\def\fighspace{-2mm}
\def\fighspacer{-2mm}
\begin{figure*}[p]
\centering
\begin{tabular}{cc}
\centering
\hspace{\fighspace}

\includegraphics[width=\figsize\textwidth]{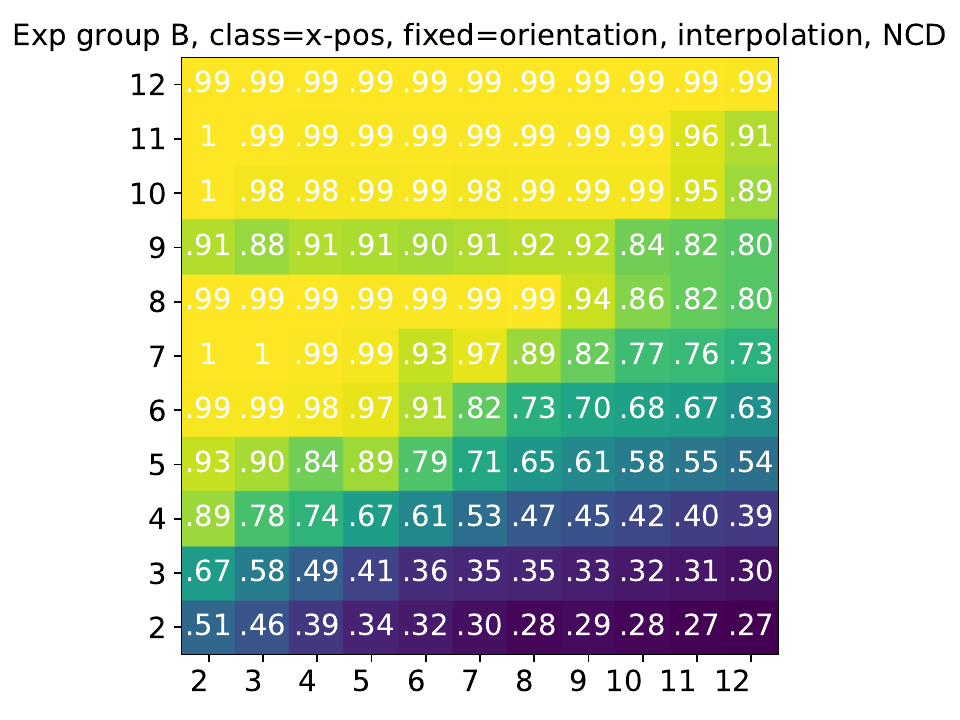} \hspace{\fighspace} & 

\includegraphics[width=\figsize\textwidth]{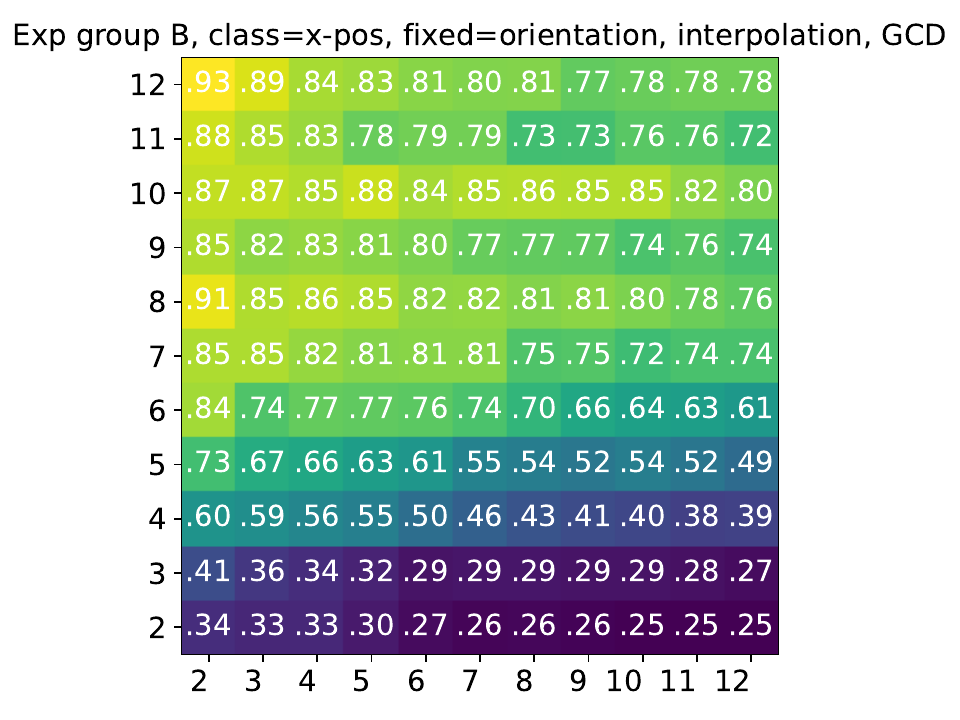}\hspace{\fighspacer}  \\

\hspace{\fighspace}\includegraphics[width=\figsize\textwidth]{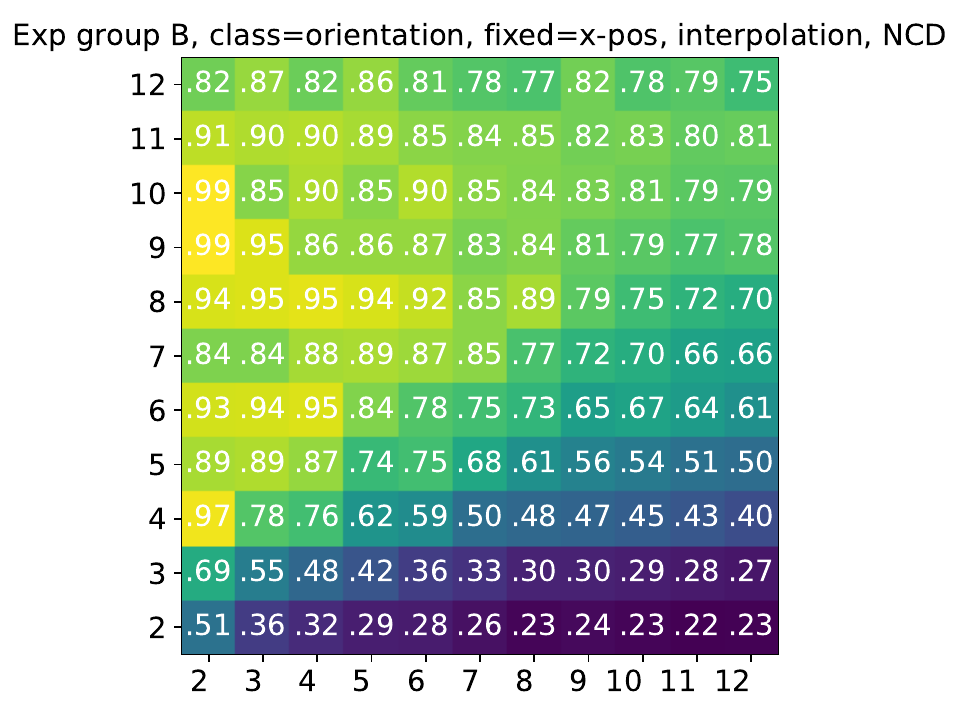}\hspace{\fighspace} & 

\includegraphics[width=\figsize\textwidth]{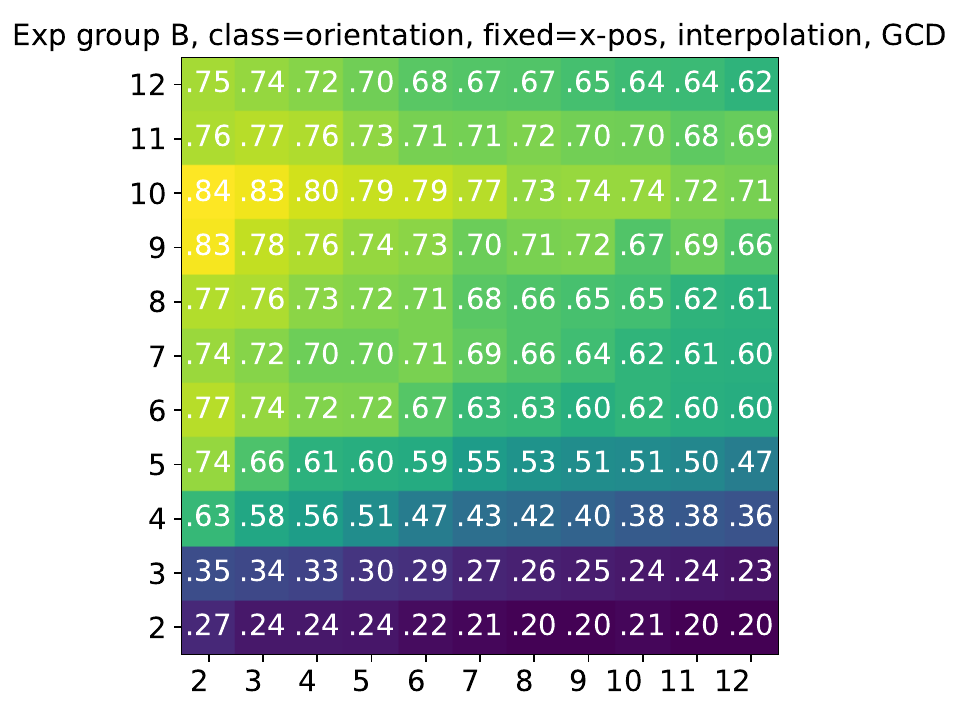}\hspace{\fighspacer} \\

\hspace{\fighspace}\includegraphics[width=\figsize\textwidth]{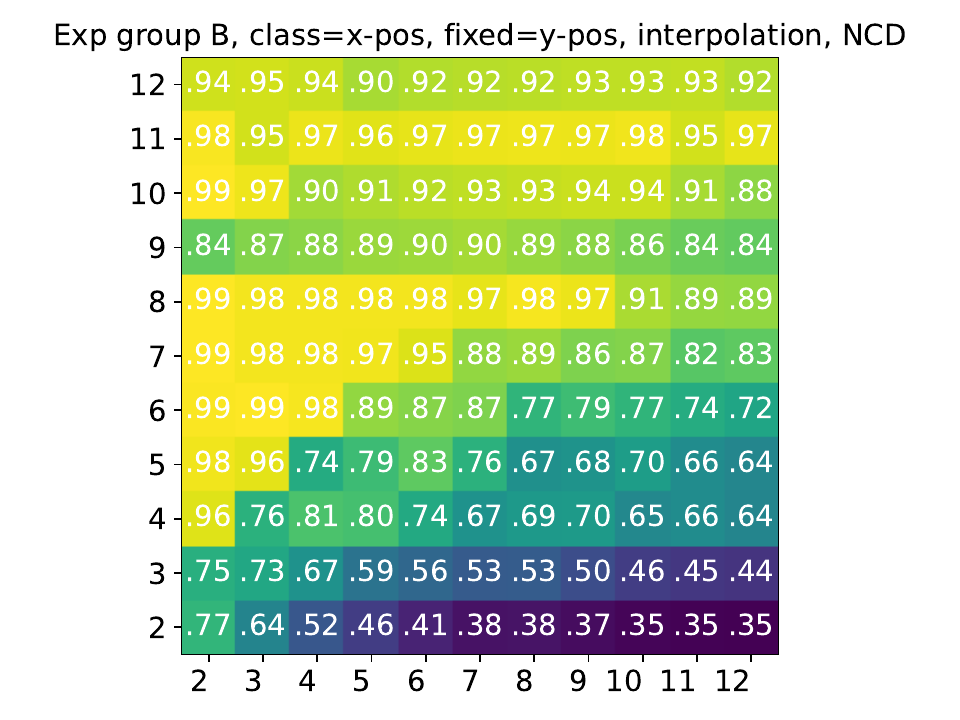}\hspace{\fighspace} & 

\includegraphics[width=\figsize\textwidth]{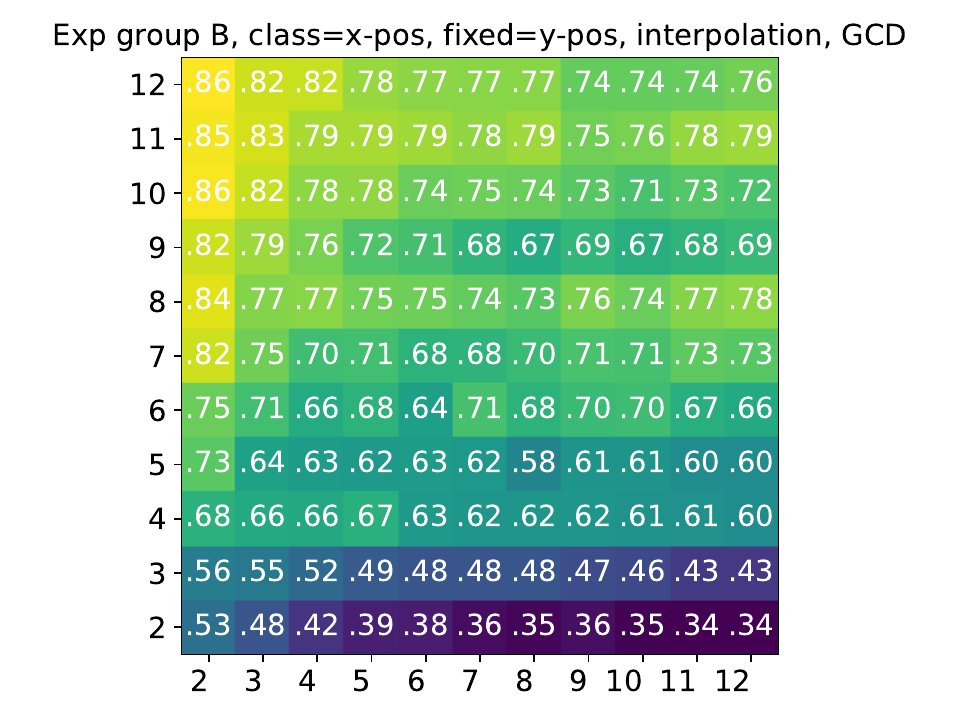}\hspace{\fighspacer} \\

\end{tabular}
\caption{ \textbf{Interpolation} results for Experiment group B : The heatmaps above are plots of the accuracy of discovering unseen classes (NCD) and classifying unseen and seen classes (GCD). This is a subgroup of the plots chosen since the other plots show similar trends.  }
\end{figure*}

\def\figsize{0.5}
\def\fighspace{-2mm}
\def\fighspacer{-2mm}
\begin{figure*}[p]
\centering
\begin{tabular}{cc}
\centering
\hspace{\fighspace}

\includegraphics[width=\figsize\textwidth]{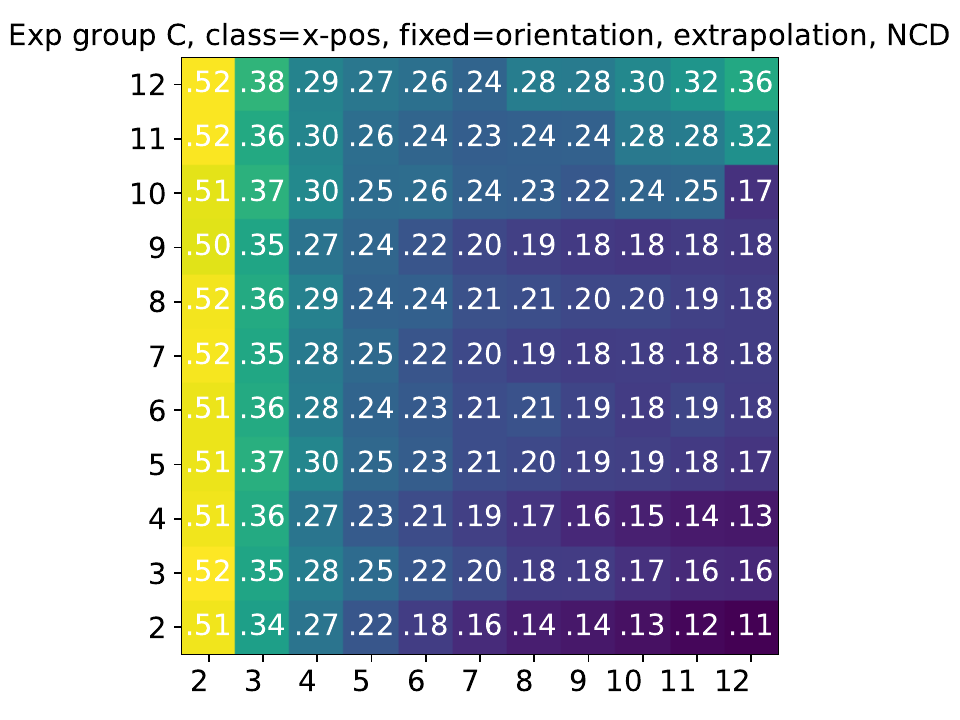} \hspace{\fighspace} & 

\includegraphics[width=\figsize\textwidth]{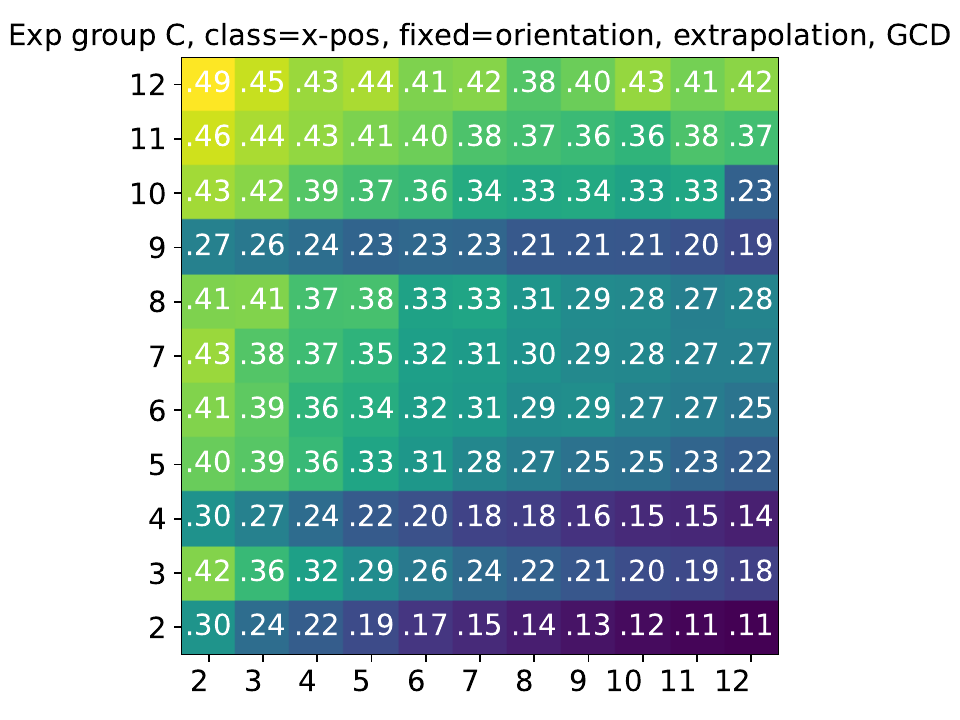}\hspace{\fighspacer}  \\

\hspace{\fighspace}\includegraphics[width=\figsize\textwidth]{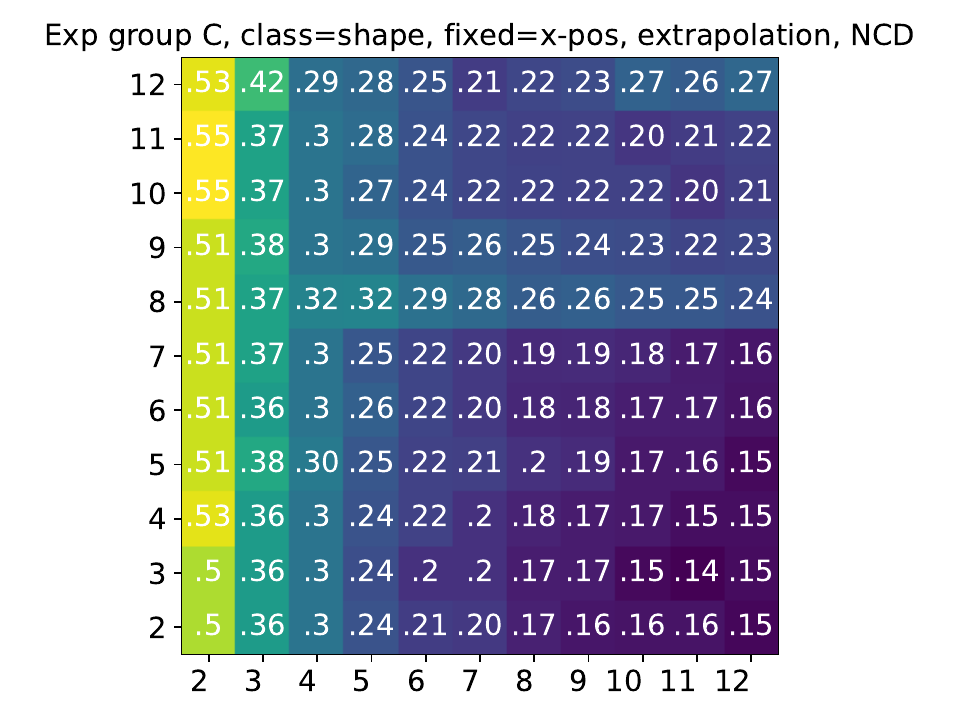}\hspace{\fighspace} & 

\includegraphics[width=\figsize\textwidth]{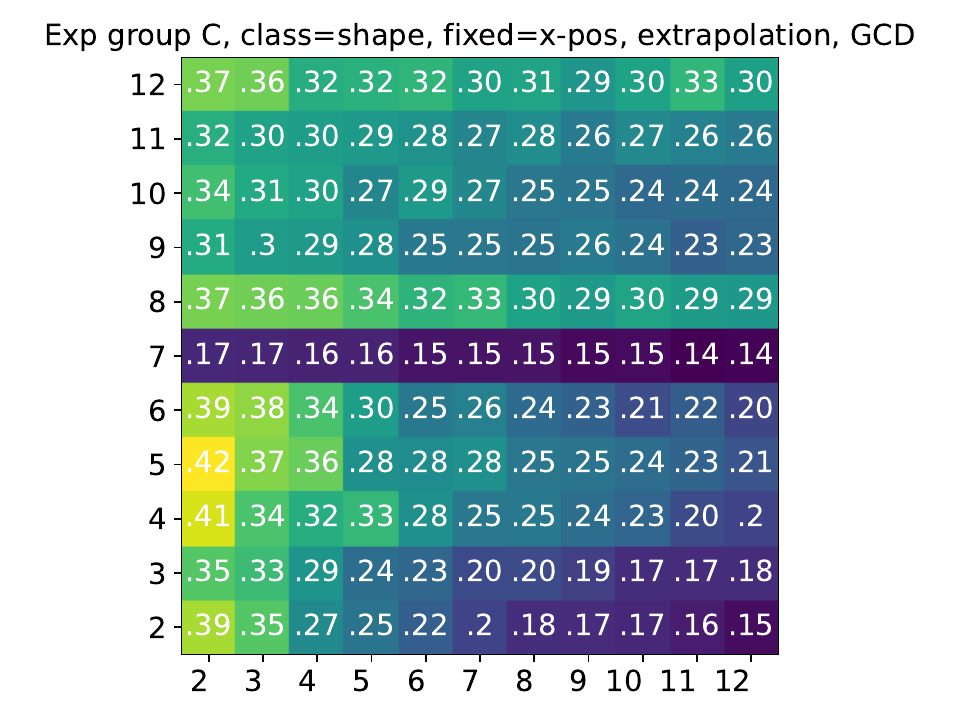}\hspace{\fighspacer} \\

\hspace{\fighspace}\includegraphics[width=\figsize\textwidth]{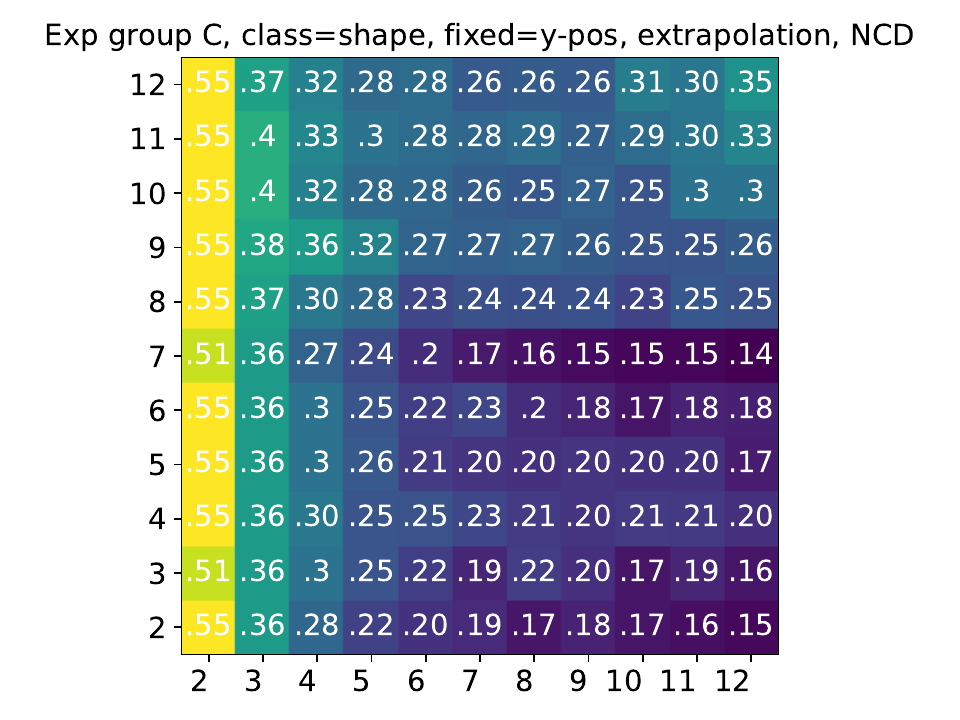}\hspace{\fighspace} & 

\includegraphics[width=\figsize\textwidth]{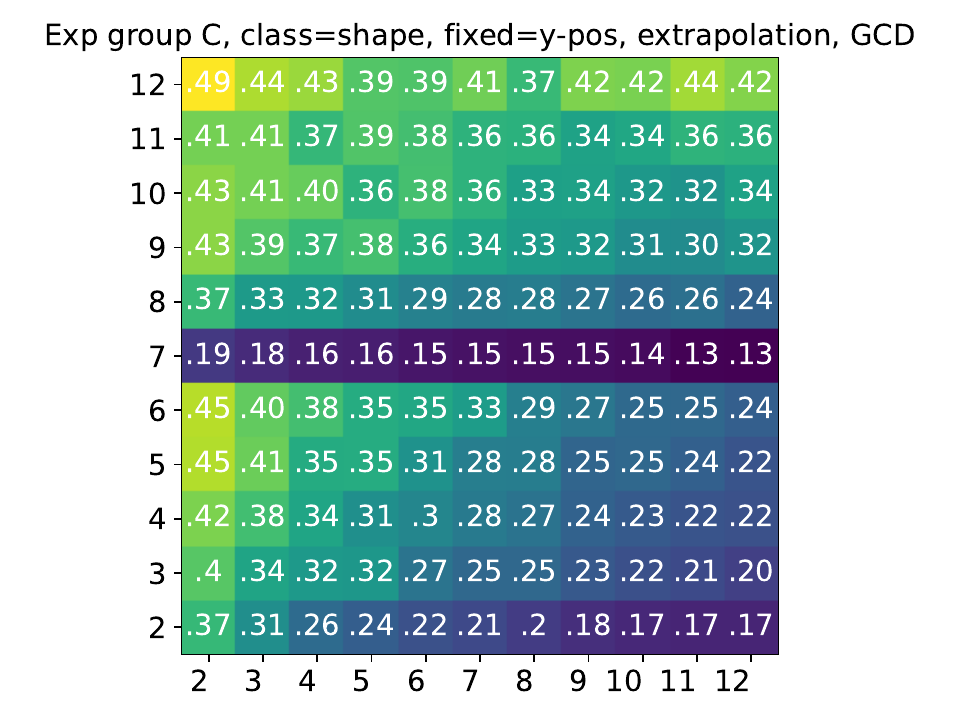}\hspace{\fighspacer} \\

\end{tabular}
\caption{ \textbf{Extrapolation} results for Experiment group C : The heatmaps above are plots of the accuracy of discovering unseen classes (NCD) and classifying unseen and seen classes (GCD).}
\end{figure*}

\def\figsize{0.5}
\def\fighspace{-2mm}
\def\fighspacer{-2mm}
\begin{figure*}[p]
\centering
\begin{tabular}{cc}
\centering
\hspace{\fighspace}

\includegraphics[width=\figsize\textwidth]{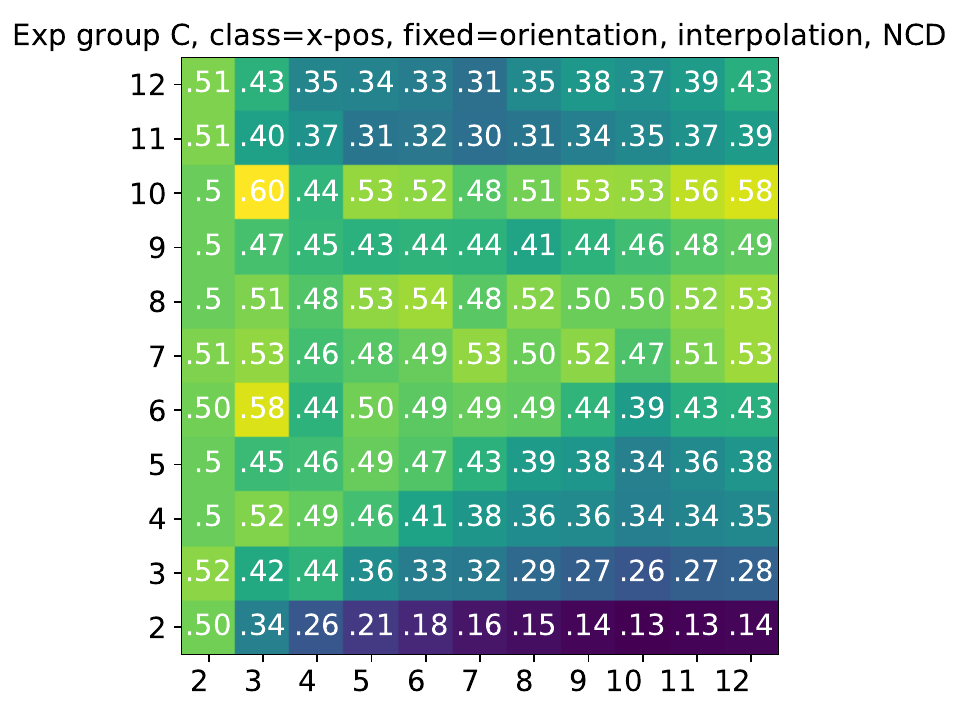} \hspace{\fighspace} & 

\includegraphics[width=\figsize\textwidth]{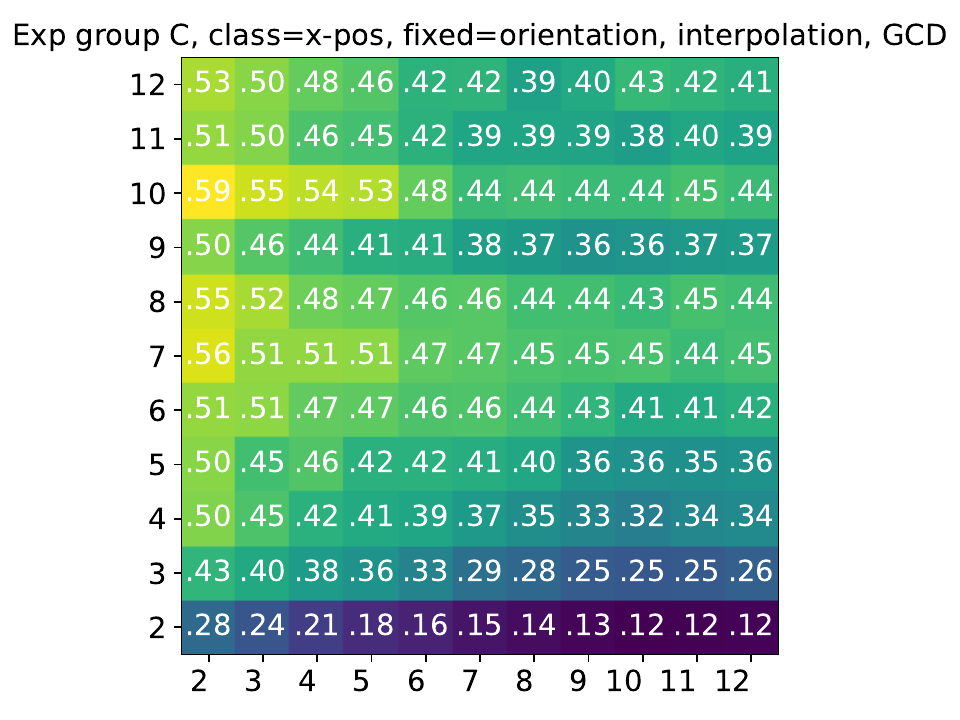}\hspace{\fighspacer}  \\

\hspace{\fighspace}\includegraphics[width=\figsize\textwidth]{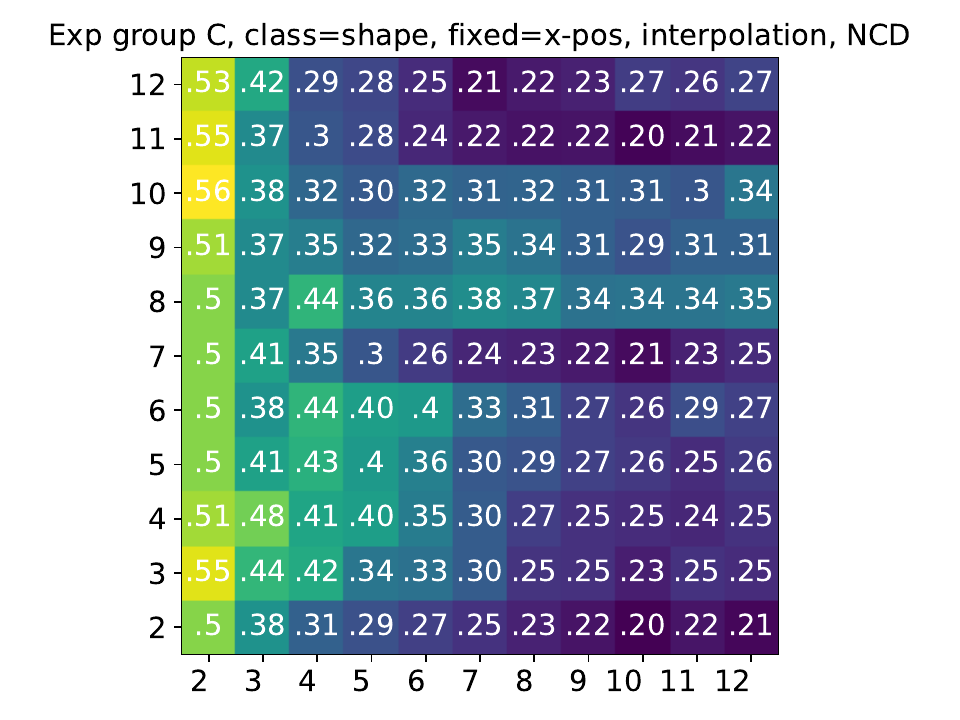}\hspace{\fighspace} & 

\includegraphics[width=\figsize\textwidth]{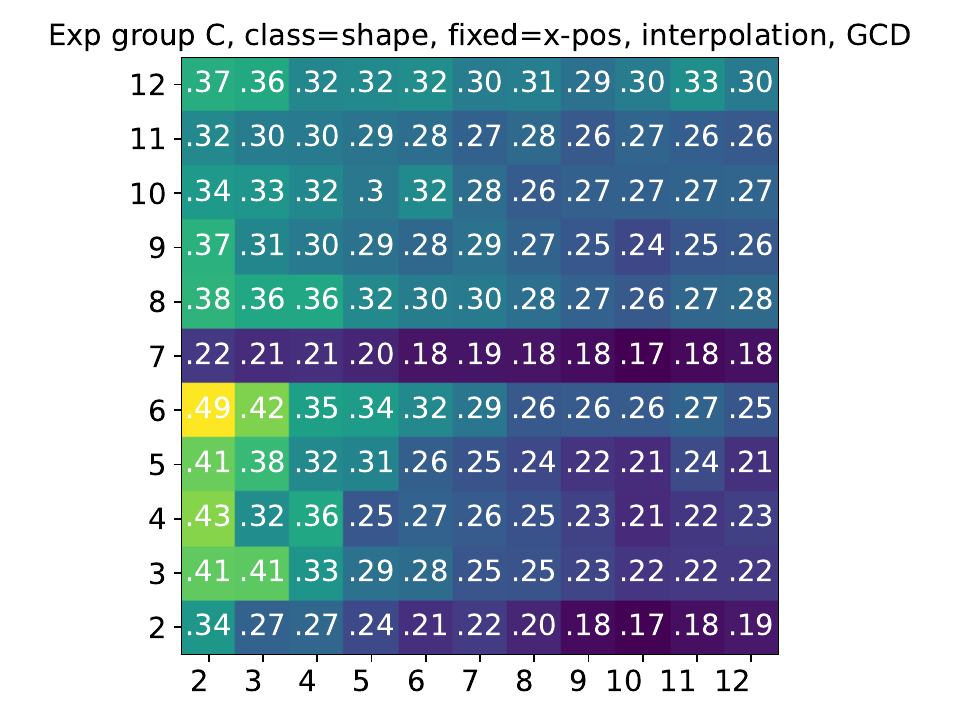}\hspace{\fighspacer} \\

\hspace{\fighspace}\includegraphics[width=\figsize\textwidth]{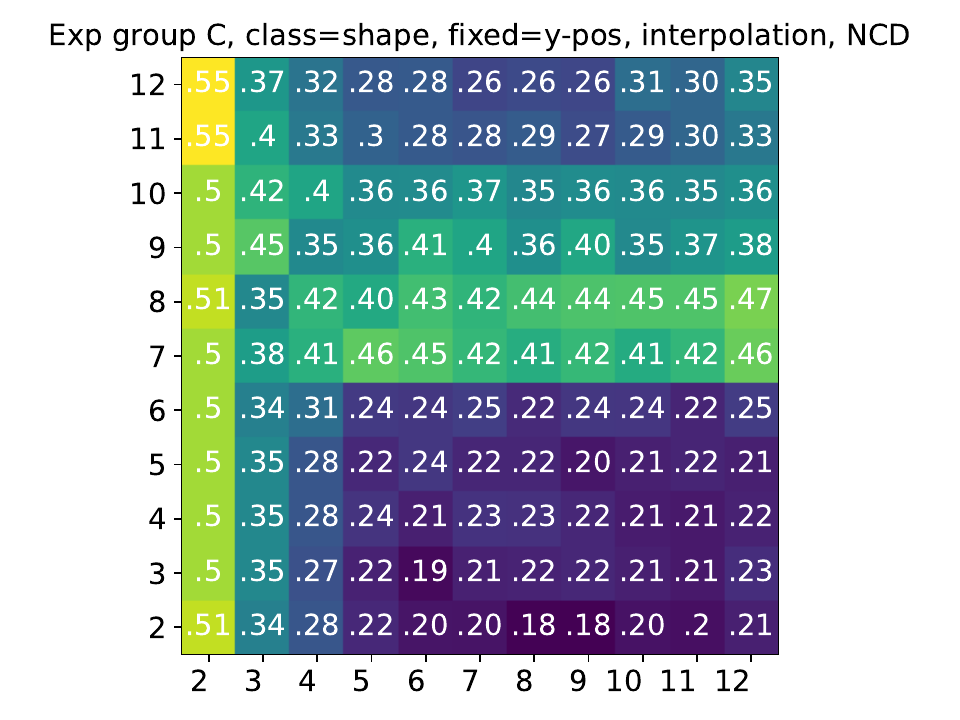}\hspace{\fighspace} & 

\includegraphics[width=\figsize\textwidth]{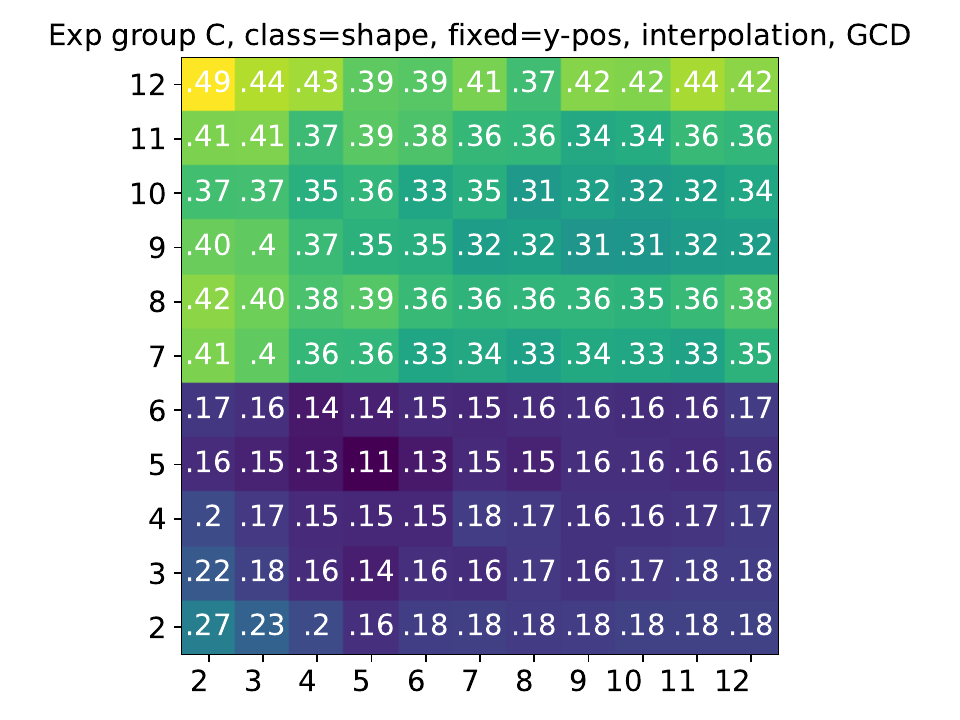}\hspace{\fighspacer} \\

\end{tabular}
\caption{ \textbf{Interpolation} results for Experiment group C : The heatmaps above are plots of the accuracy of discovering unseen classes (NCD) and classifying unseen and seen classes (GCD).}
\end{figure*}

\def\figsize{0.5}
\def\fighspace{-2mm}
\def\fighspacer{-2mm}
\begin{figure*}[p]
\centering
\begin{tabular}{cc}
\centering
\hspace{\fighspace}

\includegraphics[width=\figsize\textwidth]{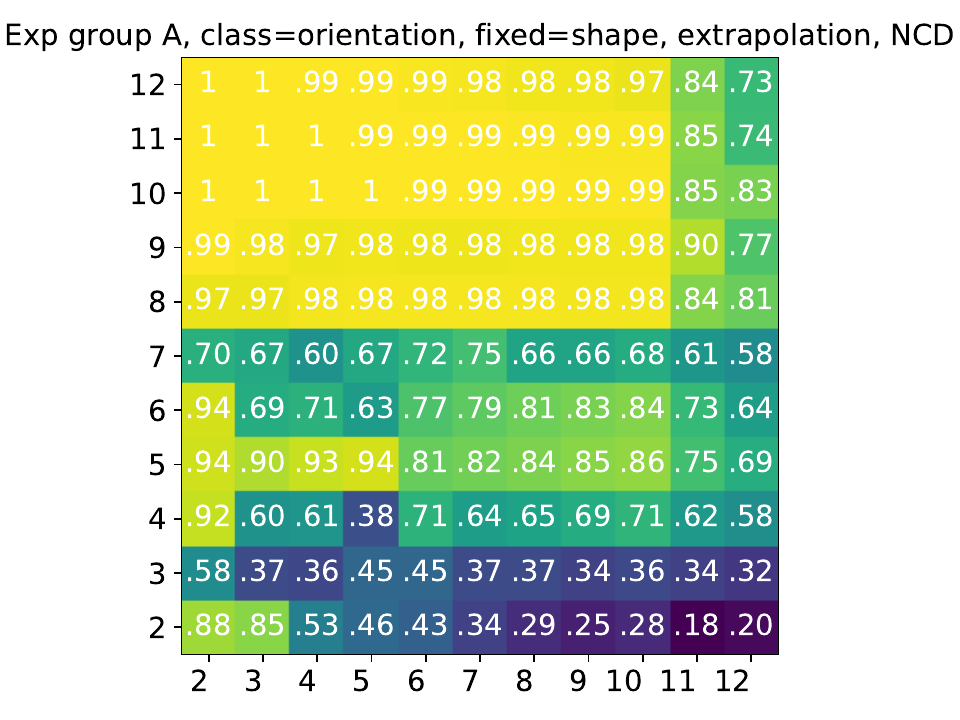} \hspace{\fighspace} & 

\includegraphics[width=\figsize\textwidth]{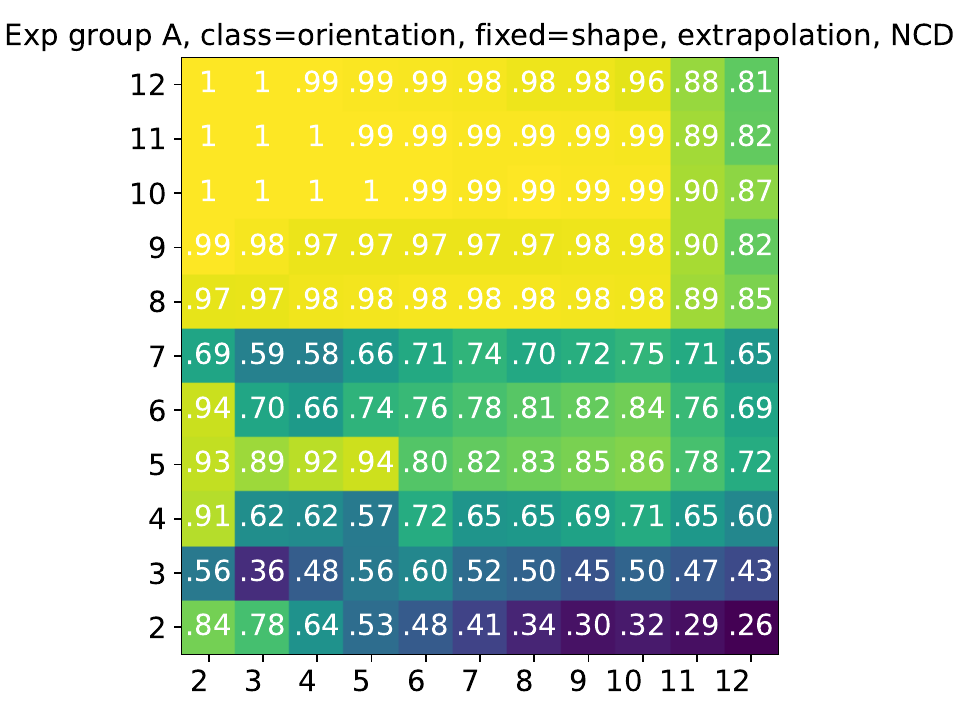}\hspace{\fighspacer}  \\

\hspace{\fighspace}\includegraphics[width=\figsize\textwidth]{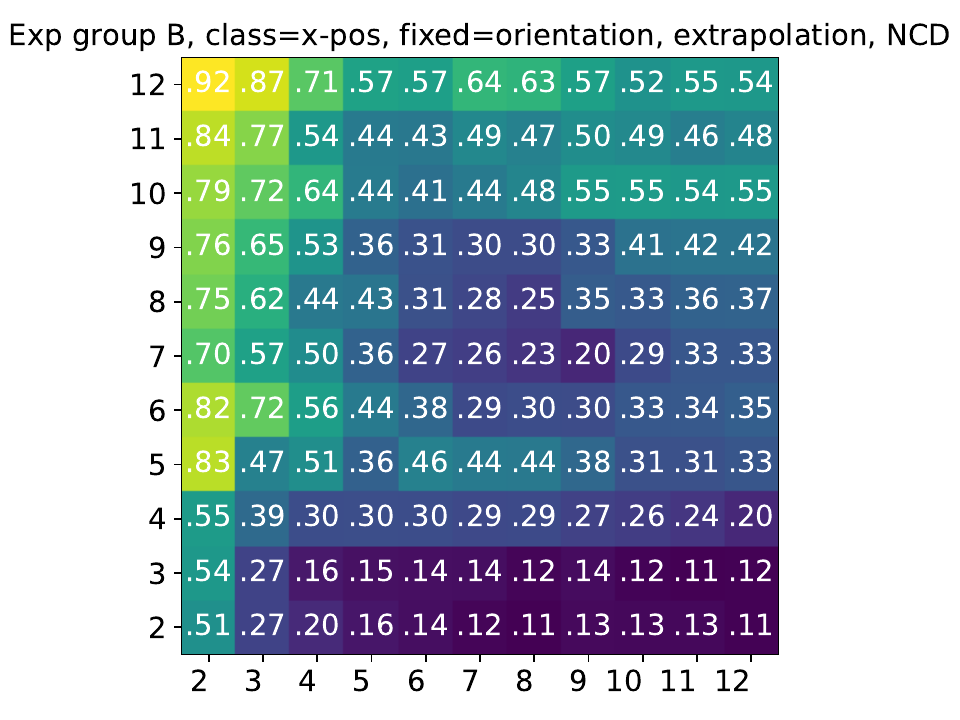}\hspace{\fighspace} & 

\includegraphics[width=\figsize\textwidth]{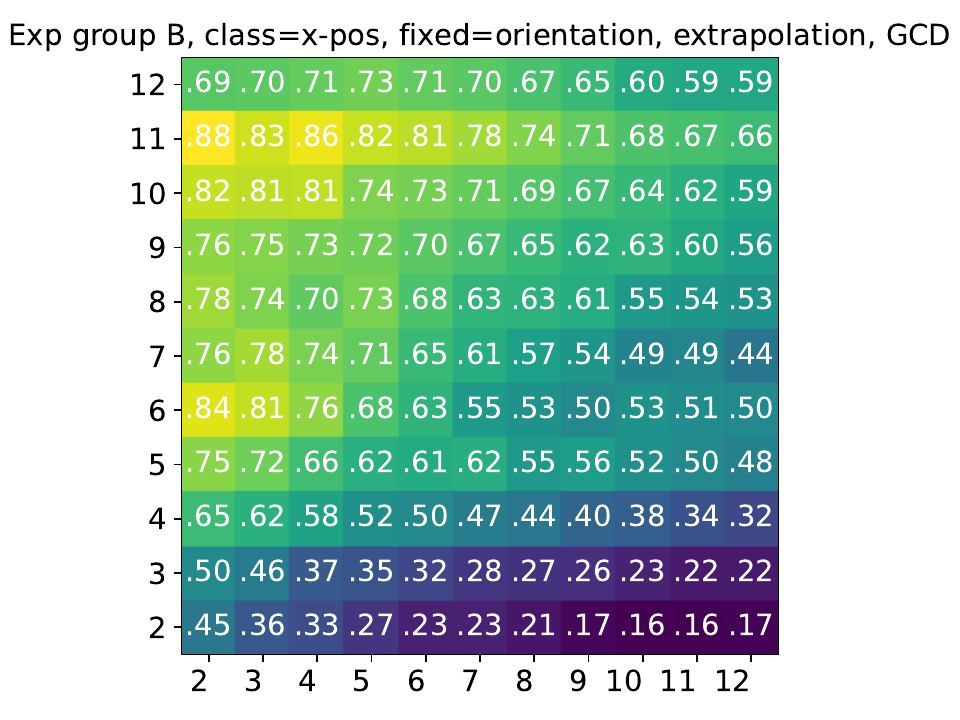}\hspace{\fighspacer} \\

\hspace{\fighspace}\includegraphics[width=\figsize\textwidth]{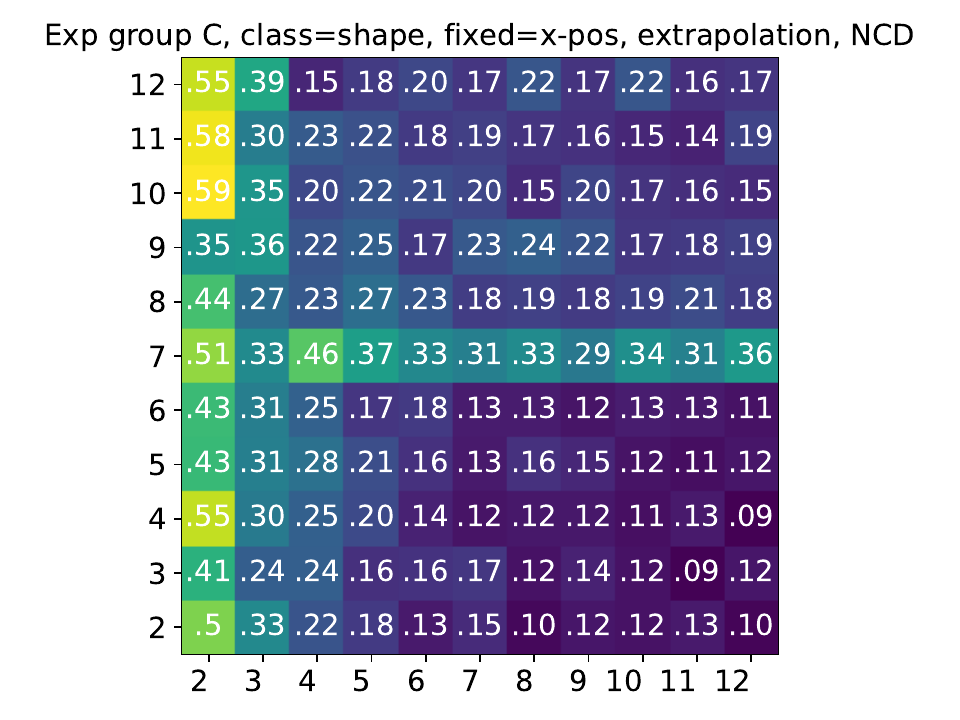}\hspace{\fighspace} & 

\includegraphics[width=\figsize\textwidth]{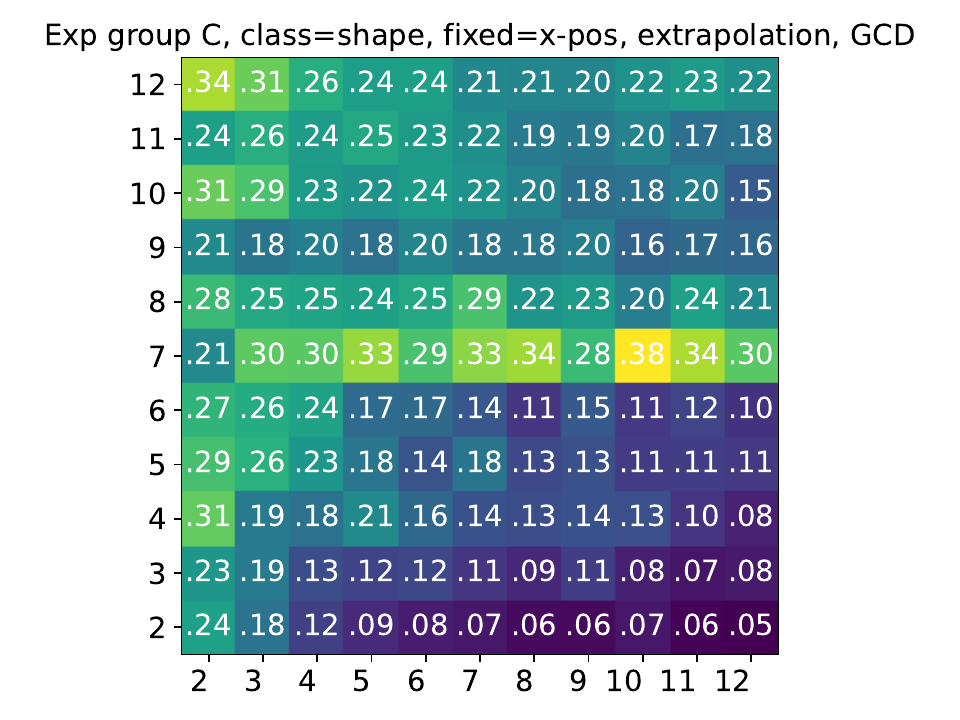}\hspace{\fighspacer} \\

\end{tabular}
\caption{ \textbf{Extrapolation} results for precision across all experiment groups. Other plots in the group show similar trends.}
\end{figure*}

\def\figsize{0.5}
\def\fighspace{-2mm}
\def\fighspacer{-2mm}
\begin{figure*}[p]
\centering
\begin{tabular}{cc}
\centering
\hspace{\fighspace}

\includegraphics[width=\figsize\textwidth]{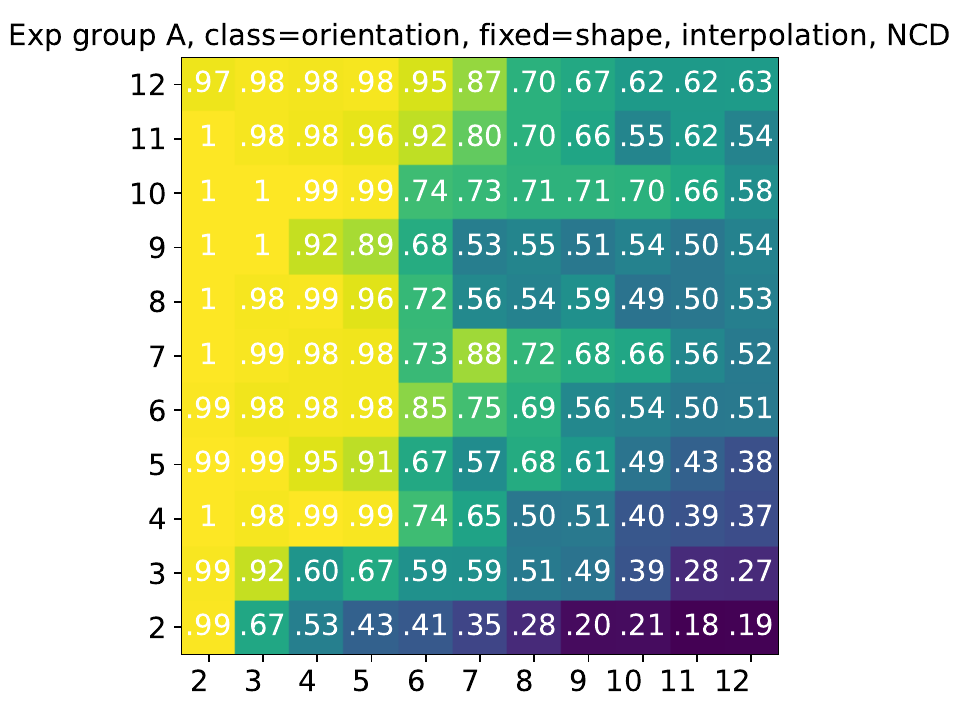} \hspace{\fighspace} & 

\includegraphics[width=\figsize\textwidth]{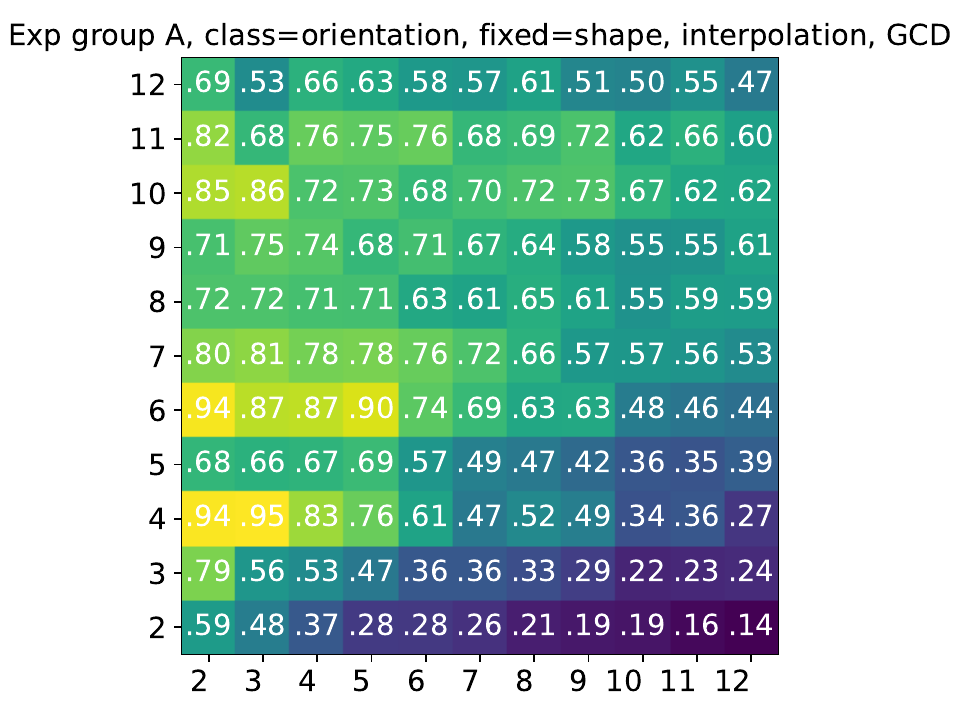}\hspace{\fighspacer} \\

\hspace{\fighspace}\includegraphics[width=\figsize\textwidth]{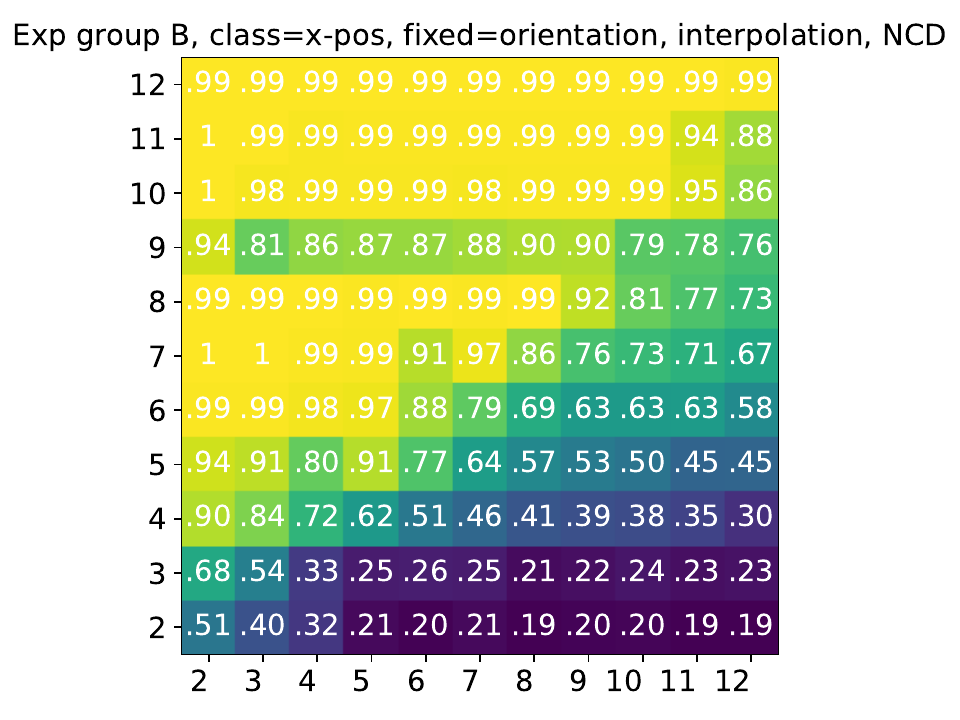}\hspace{\fighspace} & 

\includegraphics[width=\figsize\textwidth]{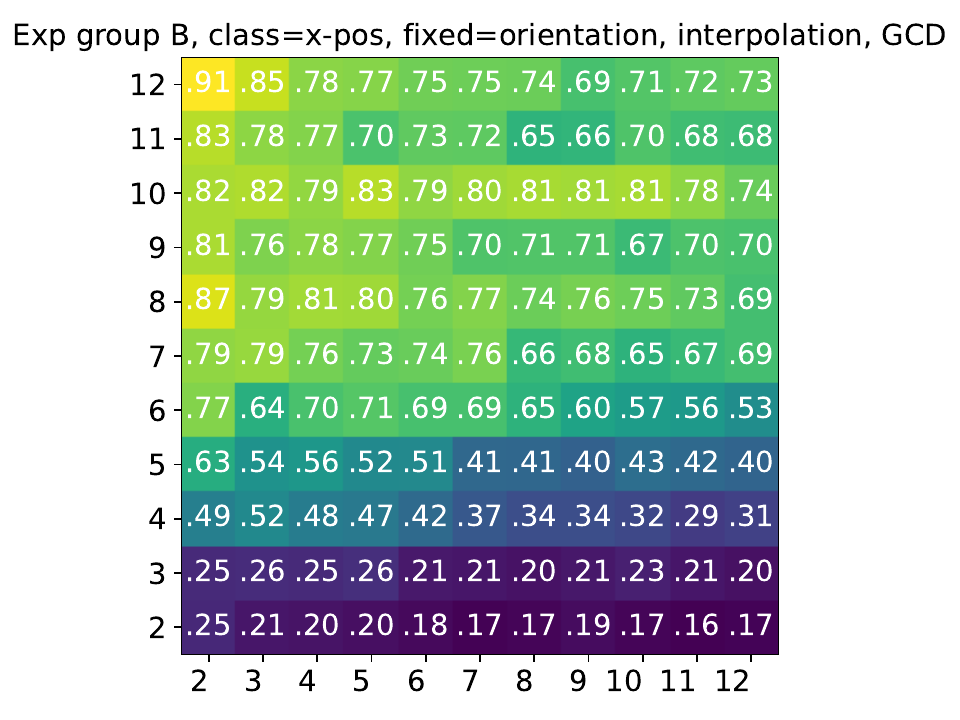}\hspace{\fighspacer} \\

\hspace{\fighspace}\includegraphics[width=\figsize\textwidth]{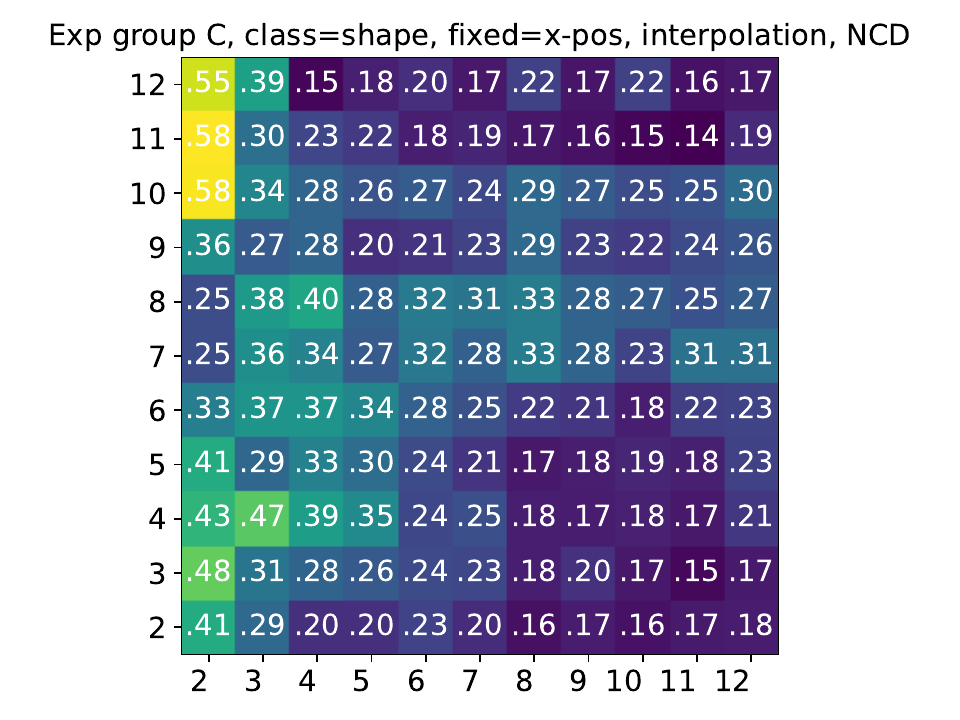}\hspace{\fighspace} & 

\includegraphics[width=\figsize\textwidth]{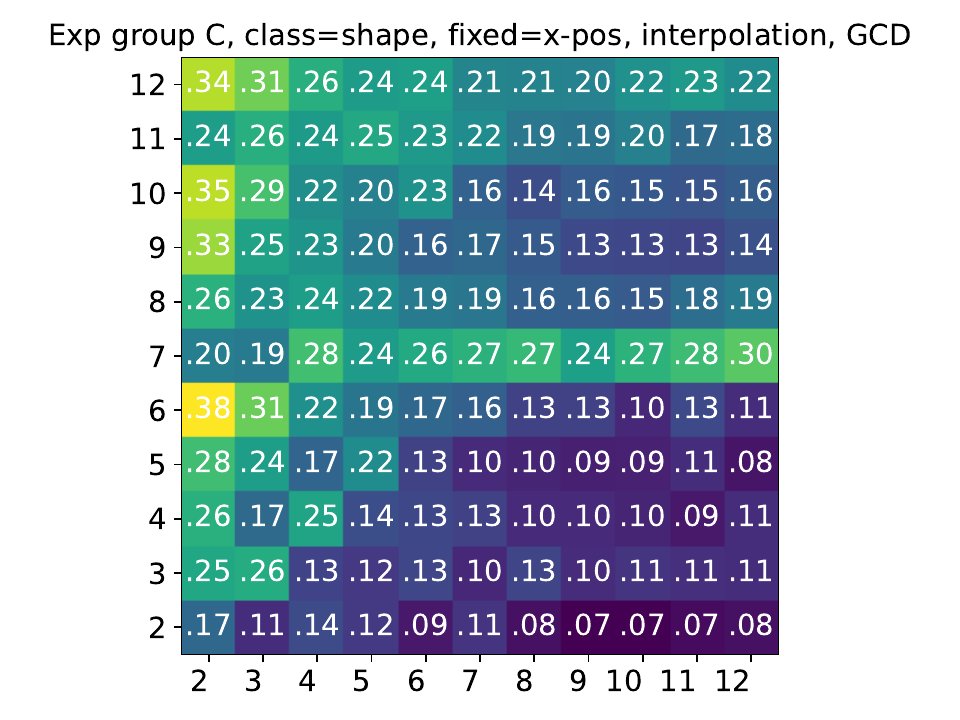}\hspace{\fighspacer} \\

\end{tabular}
\caption{ \textbf{Extrapolation} results for precision across all experiment groups. Other plots in the group show similar trends.}
\label{tab:12}
\end{figure*}

\clearpage

\section{Experiments with Model capacity}
As observed in Fig.~\ref{fig:mc_exp}, Both models exhibit a general trend of increasing accuracy as the number of parameters per known class increases. This suggests that larger model capacities relative to the number of known classes help in achieving better feature representation, leading to improved class discovery. However, the improvements tend to diminish at higher parameter-to-class ratios, suggesting that beyond a certain capacity, the additional model complexity does not significantly enhance discovery performance. This finding is consistent with the idea that the learning of novel class representations is constrained by factors such as feature diversity in the known classes and the inherent variability in the unknown classes.

There is no consistent trend where ResNet-50 (red) significantly outperforms ResNet-18 (blue). In many cases, the performances overlap, particularly at lower parameter-to-class ratios. Therefore, increase in model capacity does not significantly help the discovery task.

\section{Experiments on Real Dataset (CUB-200 Dataset)} 

We conduct experiments on the CUB-200 dataset to demonstrate how the trends observed in synthetic experiments generalize to real-world data. The heat maps ~\ref{tab:bird_heat} for accuracy and precision are included to showcase that performance varies depending on the number of known and unknown classes, with higher accuracy and precision generally achieved when the number of known classes is larger and the number of unknown classes is smaller.

{\bf Interpolation and extrapolation}. The heatmaps (Figure~\ref{tab:bird_heat}) reveal no clear general trend in the difference between extrapolation and interpolation across precision and accuracy metrics. The variations in differences appear context-dependent, with both positive (red) and negative (blue) values scattered across the range of known and unknown classes without a consistent pattern. While certain configurations tend to perform better, there is no universal trend. Similarly, interpolation does not consistently outperform extrapolation in scenarios with many unknown classes.

We also plot the relative change in accuracy and precision in Fig. ~\ref{tab:change} based on the number of known and unknown classes. Both accuracy and precision exhibit a saturation effect after the number of unknown classes exceeds approximately four. This suggests that the primary challenge in discovering novel classes arises early, and adding more unknown classes beyond this point does not substantially degrade performance. The figure indicates that the NCD system stabilizes as the number of unknown classes increases, with performance metrics leveling off following the initial drop.

\def\figsize{0.6}
\def\fighspace{-15mm}
\def\fighspacer{100mm}
\begin{figure*}[p]
\hspace*{-10mm}
\begin{tabular}{cc}
\centering
\hspace{\fighspace}

\includegraphics[width= 0.65\textwidth]{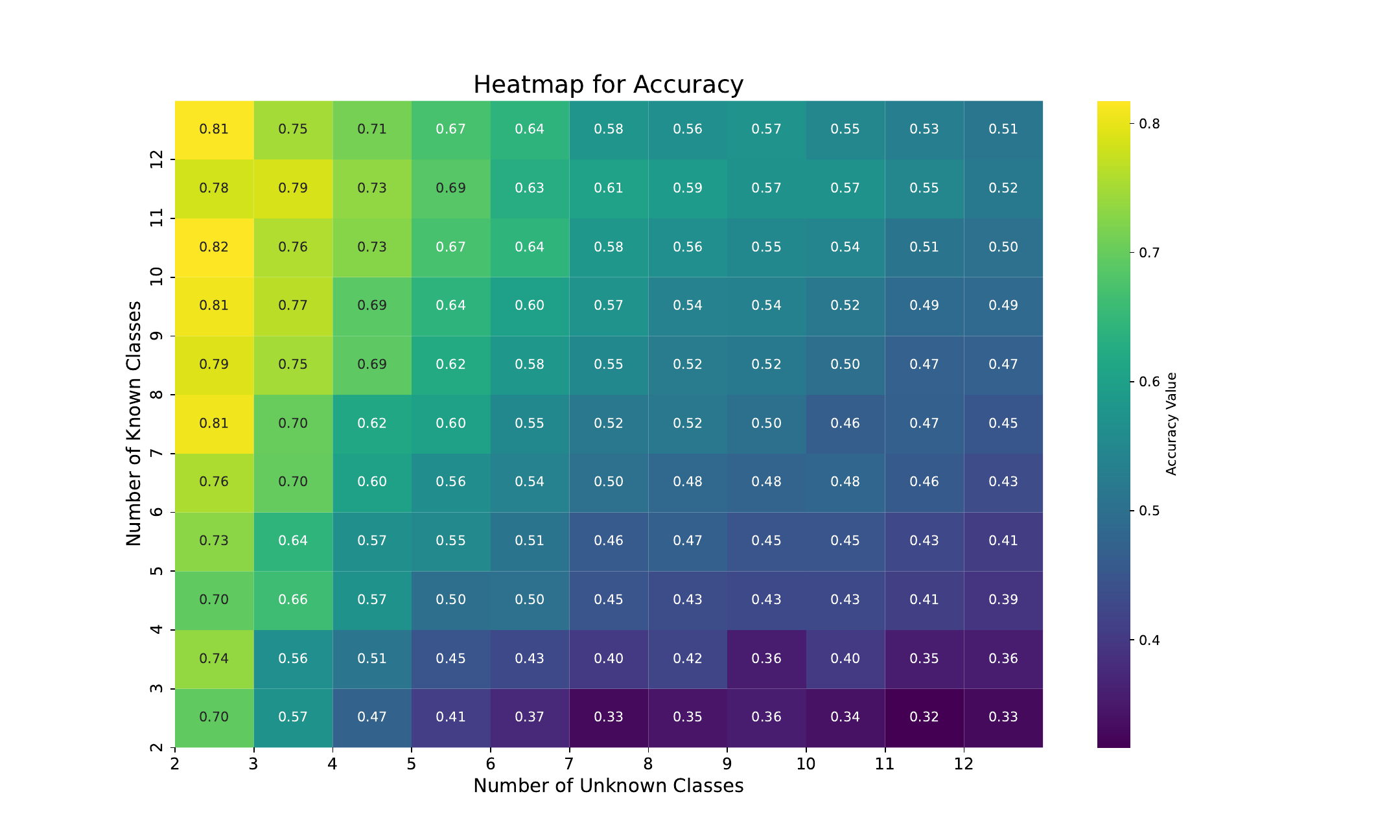}
\hspace{\fighspace} &

\includegraphics[width= 0.65\textwidth]{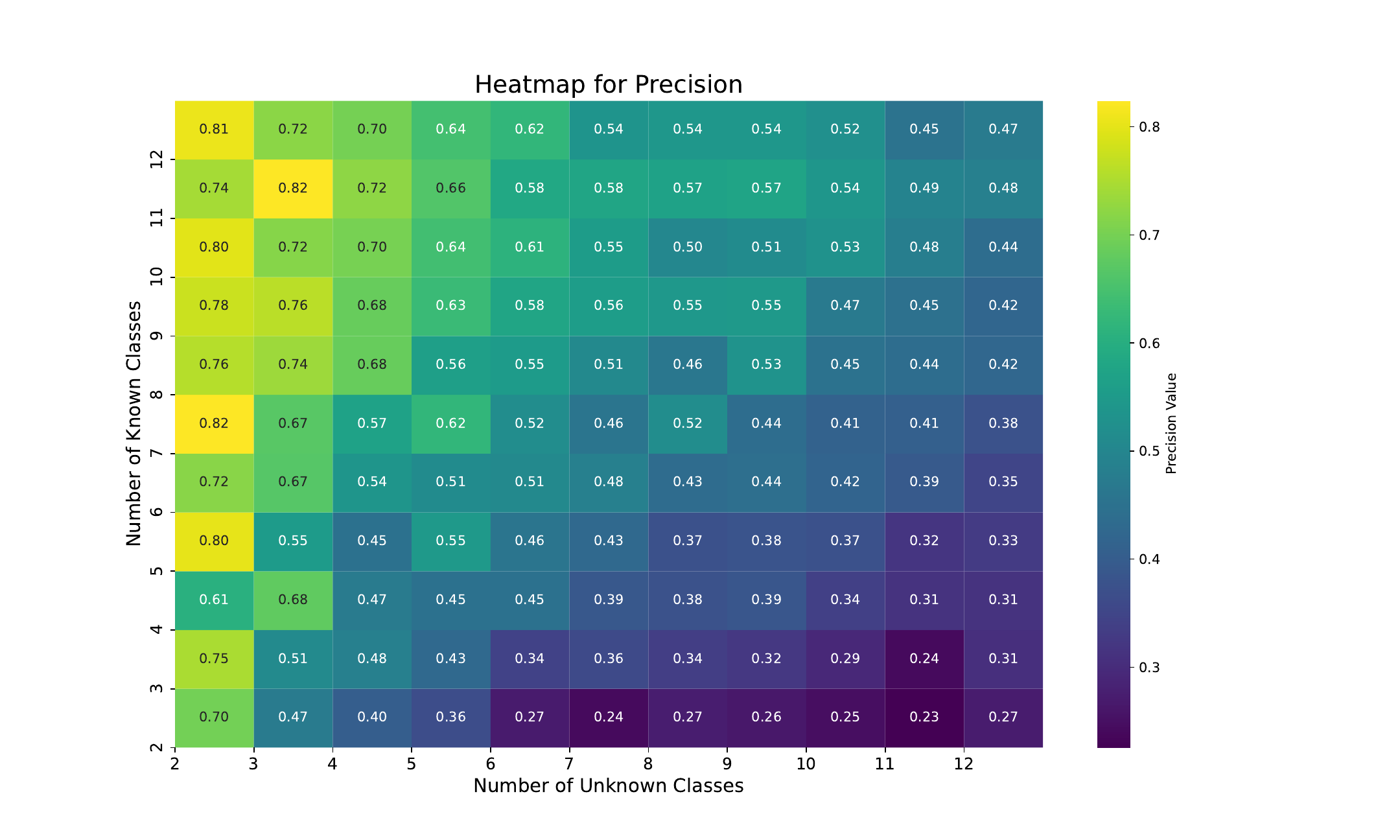}  \hspace{\fighspacer} 
\label{fig:fig6}
\end{tabular}

\caption{Results from CUB-200 dataset experiment with row as number of known and column as unknown.}
\label{tab:bird_heat}

\centering
\begin{tabular}{cc}
\centering
\hspace{\fighspace}

\includegraphics[width=\figsize\textwidth]{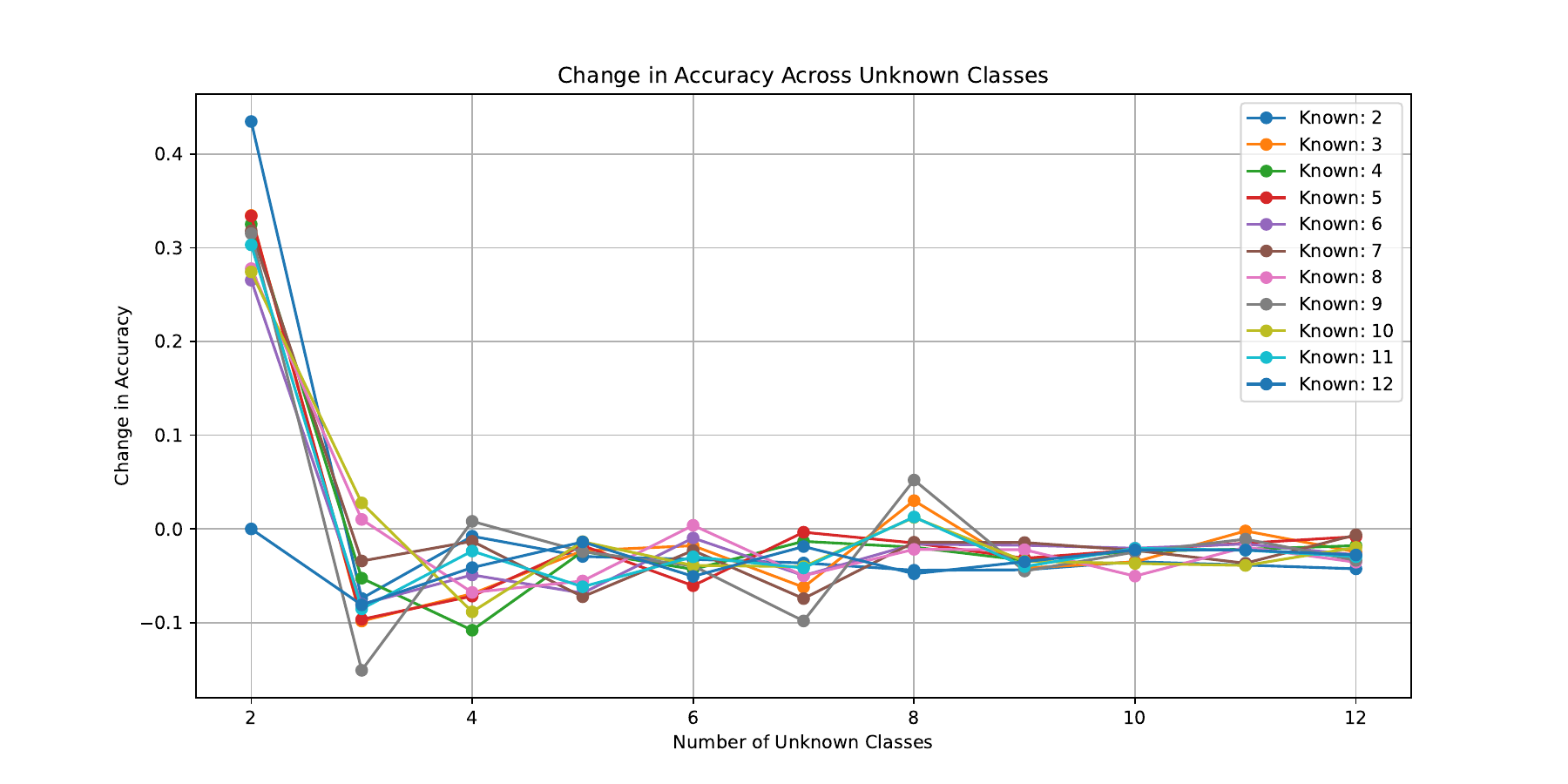}
\hspace{\fighspace} &

\includegraphics[width=\figsize\textwidth]{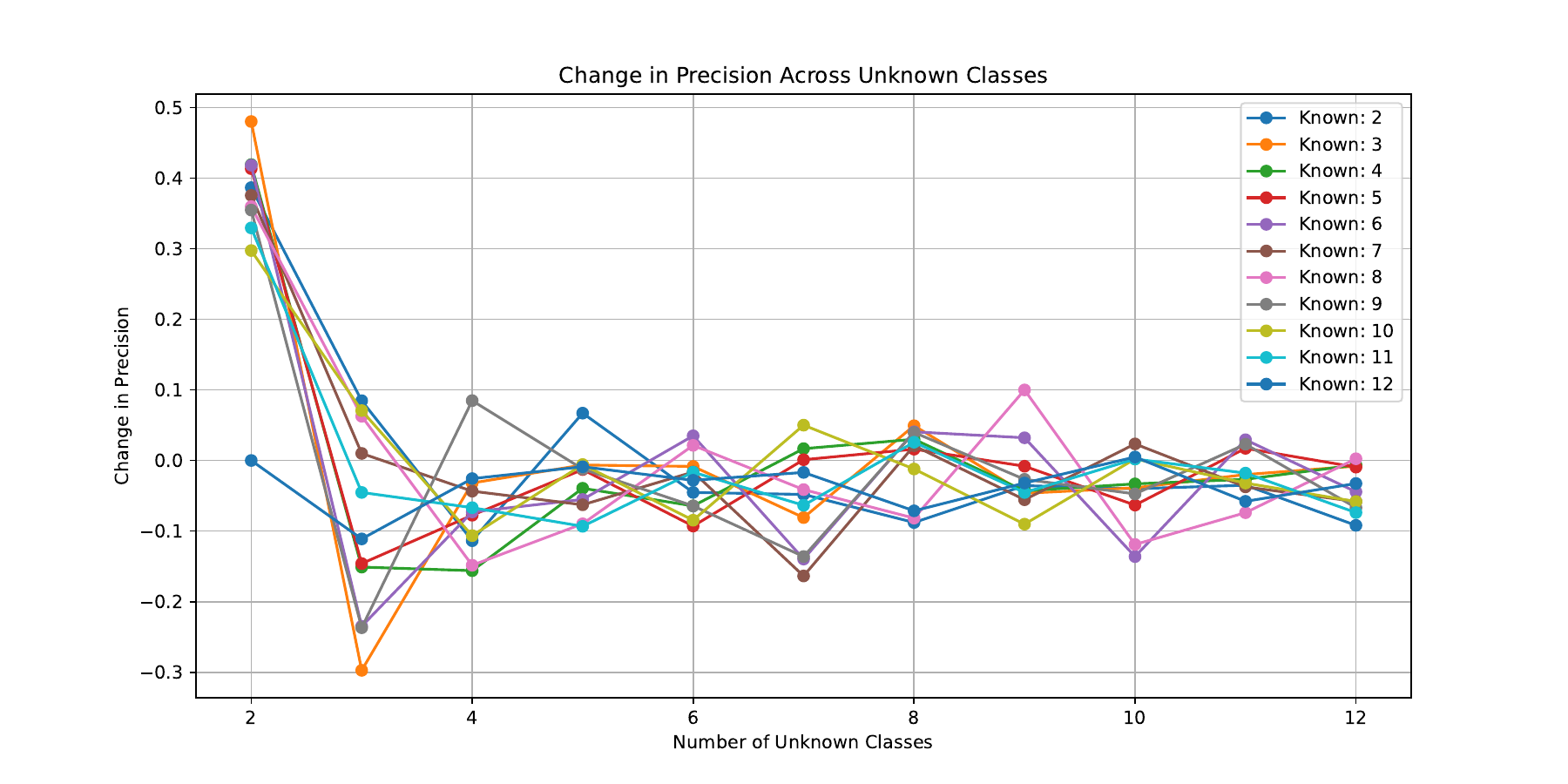}  \hspace{\fighspacer} 
\label{fig:fig6}
\end{tabular}
\caption{Plotting the change in metrics to show existence of saturation point.}
\label{tab:change}

\hspace*{-2cm}
\begin{tabular}{cc}
\centering
\def\fighspace{-35mm}
\def\fighspacer{-100mm}
\includegraphics[width=0.7\textwidth]{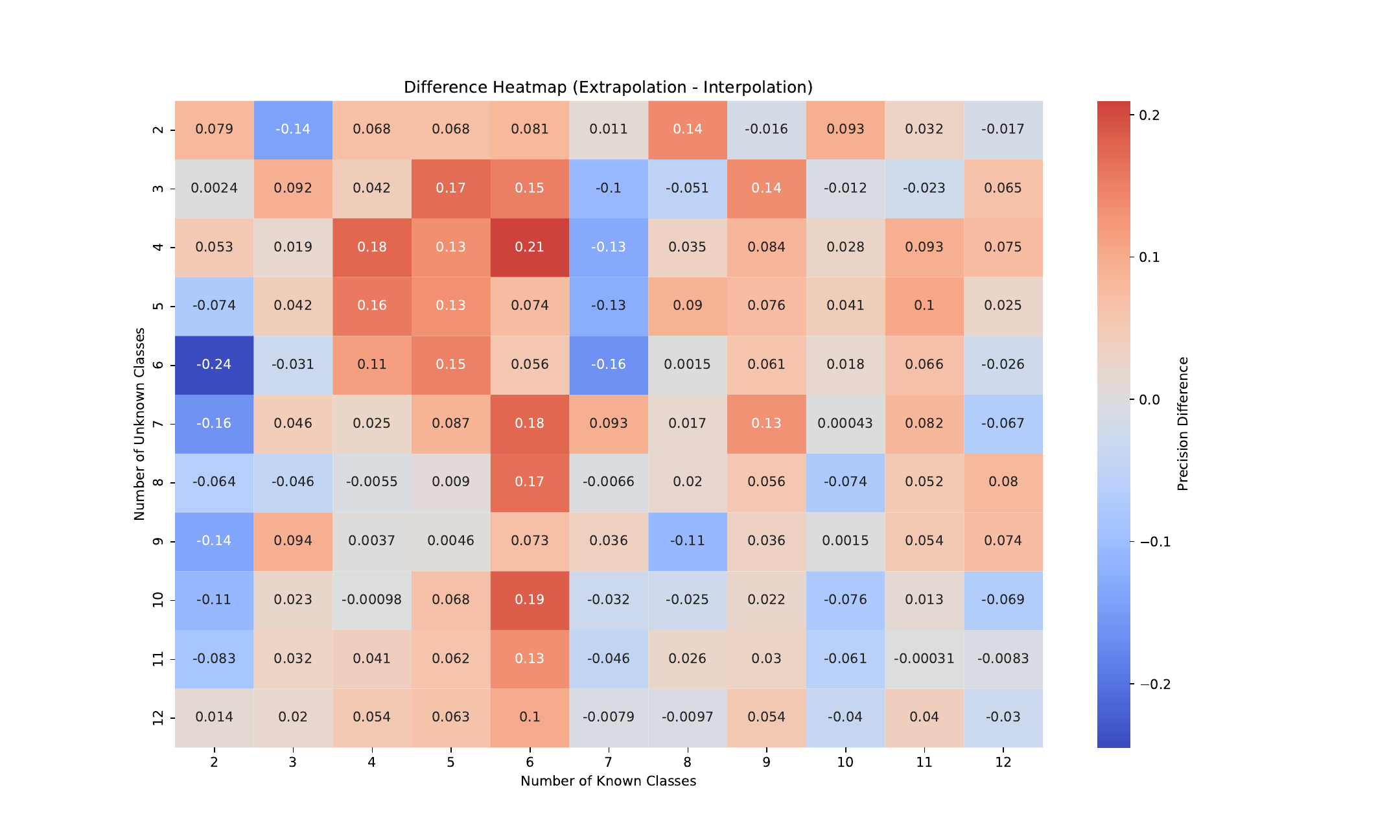} \hspace{\fighspace} & 

\includegraphics[width=0.7\textwidth]{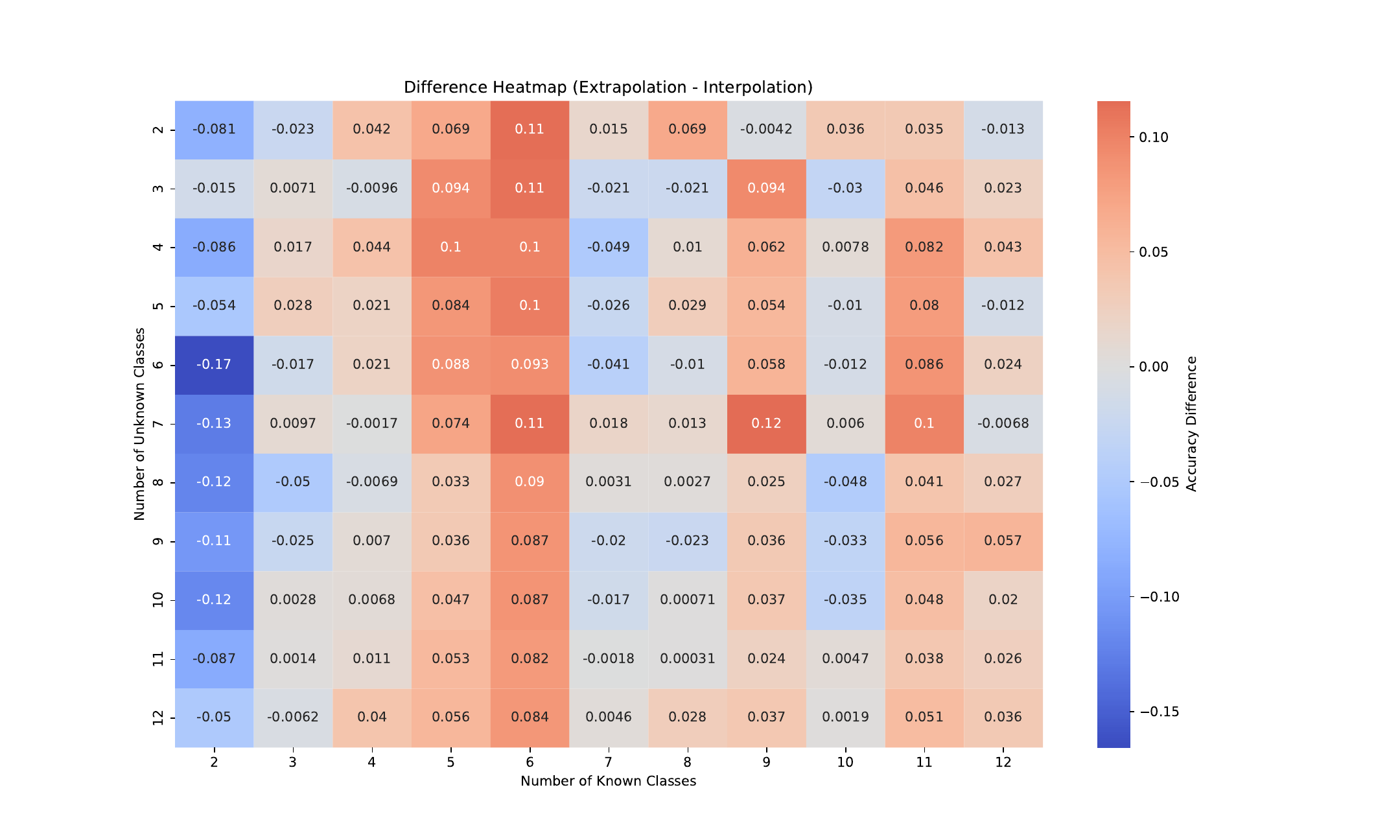} \hspace{\fighspacer} 
\end{tabular}
\caption{Heatmap showing the difference in Extrapolation and Interpolation cases performed with CUB-200 dataset.}
\end{figure*}

\def\figsize{0.6}
\def\fighspace{-15mm}
\def\fighspacer{200mm}
\begin{figure*}[h]
\centering
\begin{tabular}{cc}
\centering
\hspace{\fighspace}

\includegraphics[width=\figsize\textwidth]{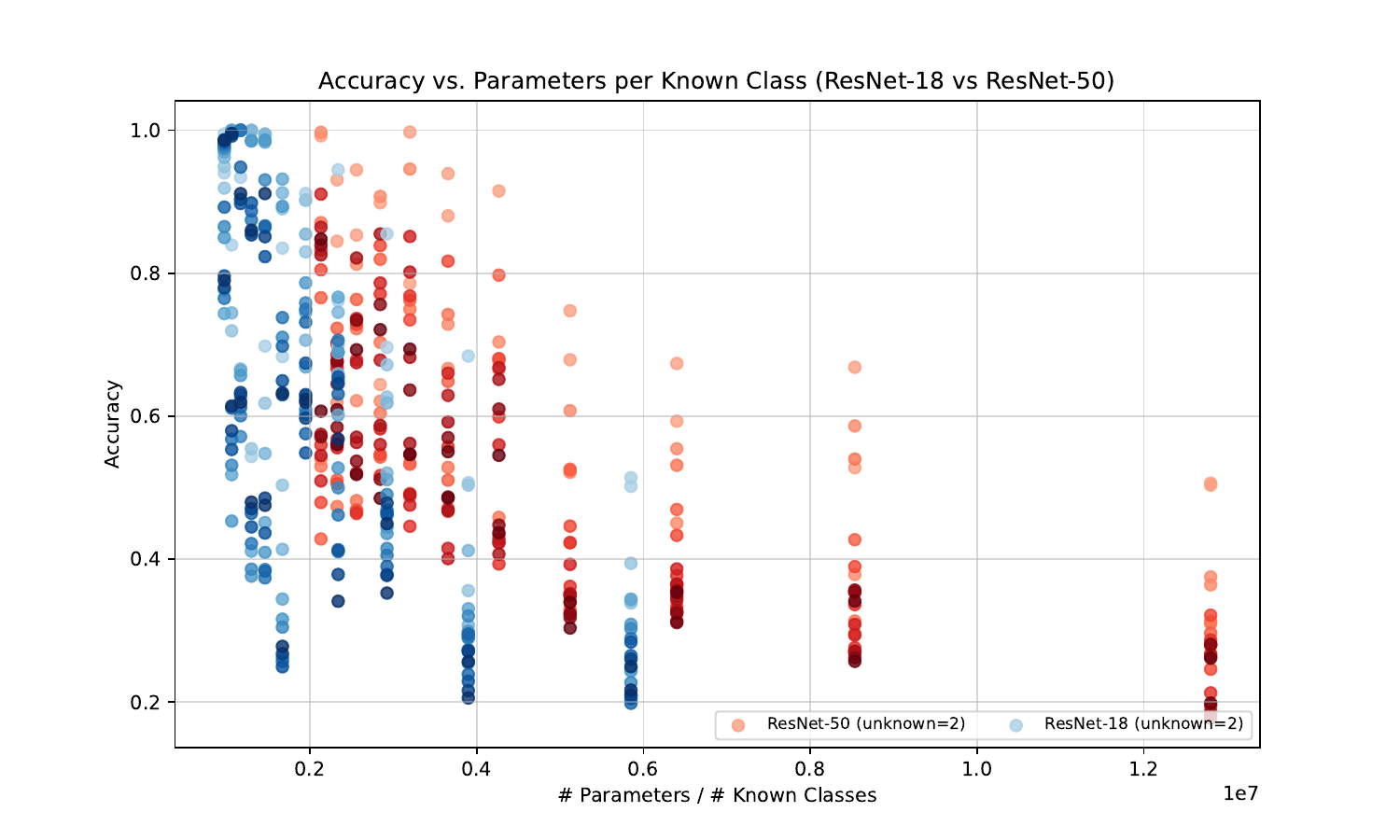} \hspace{\fighspace} &

\includegraphics[width=\figsize\textwidth]{figures/model_capacity/mc_accuracy.pdf} \hspace{\fighspacer}
\end{tabular}
\caption{\small This figure illustrates the impact of model capacity (measured as the number of parameters divided by the number of known classes) on the accuracy of novel class discovery for ResNet-18 and ResNet-50. The x-axis represents the number of parameters per known class, while the y-axis shows the accuracy achieved in clustering-based discovery of unknown classes. ResNet-18 results are displayed in shades of blue, and ResNet-50 results are in shades of red. For each known class configuration, different shades of the same base color represent varying numbers of unknown classes, representing how increasing the number of unknown classes impacts performance. The plots were made across Experiment A and B all of which exhibit similar trends.}
\label{fig:mc_exp}
\end{figure*}

\clearpage



\end{document}